\pdfoutput=1
\documentclass[11pt, a4]{article}

\usepackage[]{naacl2021}

\usepackage{times}
\usepackage{latexsym}

\usepackage[T1]{fontenc}
\usepackage[utf8]{inputenc}

\usepackage{microtype}

\usepackage{booktabs,graphicx}

\title{Dynabench: Rethinking Benchmarking in NLP}
\author{
Douwe Kiela$^\dagger$, Max Bartolo$^\ddagger$, Yixin Nie$^\star$, Divyansh Kaushik$^\mathsection$, Atticus Geiger$^\mathparagraph$, 
\AND
Zhengxuan Wu$^\mathparagraph$, Bertie Vidgen$^\|$, Grusha Prasad$^{\star\star}$, Amanpreet Singh$^\dagger$, Pratik Ringshia$^\dagger$,
\AND
Zhiyi Ma$^\dagger$, Tristan Thrush$^\dagger$, Sebastian Riedel$^{\dagger\ddagger}$, Zeerak Waseem$^{\dagger\dagger}$, Pontus Stenetorp$^\ddagger$,
\AND
Robin Jia$^\dagger$, Mohit Bansal$^\star$, Christopher Potts$^\mathparagraph$ and Adina Williams$^\dagger$\\[1ex]
$^\dagger$ Facebook AI Research; $^\ddagger$ UCL; $^\star$ UNC Chapel Hill; $^\mathsection$ CMU; $^\mathparagraph$ Stanford University\\
$^\|$ Alan Turing Institute; $^{\star\star}$ JHU; $^{\dagger\dagger}$ Simon Fraser University\\
  \texttt{dynabench@fb.com}\\}

\begin{document}
\maketitle
\begin{abstract}
We introduce Dynabench, an open-source platform for dynamic dataset creation and model benchmarking. Dynabench runs in a web browser and supports human-and-model-in-the-loop dataset creation: annotators seek to create examples that a target model will misclassify, but that another person will not. In this paper, we argue that Dynabench addresses a critical need in our community: contemporary models quickly achieve outstanding performance on benchmark tasks but nonetheless fail on simple challenge examples and falter in real-world scenarios. 
With Dynabench, dataset creation, model development, and model assessment can directly inform each other, leading to more robust and informative benchmarks. We report on four initial NLP tasks, illustrating these concepts and highlighting the promise of the platform, and address potential objections to dynamic benchmarking as a new standard for the field.
\end{abstract}

\section{Introduction}

%While it took roughly 20 years for machine learning models to surpass estimates of human performance on Switchboard conversational speech recognition \citep{Godfrey:Holliman:1997,xiong2017the}, 

%While it took roughly 18 years for machine learning models to surpass estimates of human performance on MNIST~\cite{Lecun1998gradient,Cirecsan2012multi,Wan2013dropconnect}, that same milestone was reached in less than a year for the GLUE benchmark~\citep{wang-etal-2018-glue}. It feels like our field has made remarkable progress in a very short time span.

\begin{figure}[t]
    \centering\vspace{10pt}
    \includegraphics[width=0.98\columnwidth]{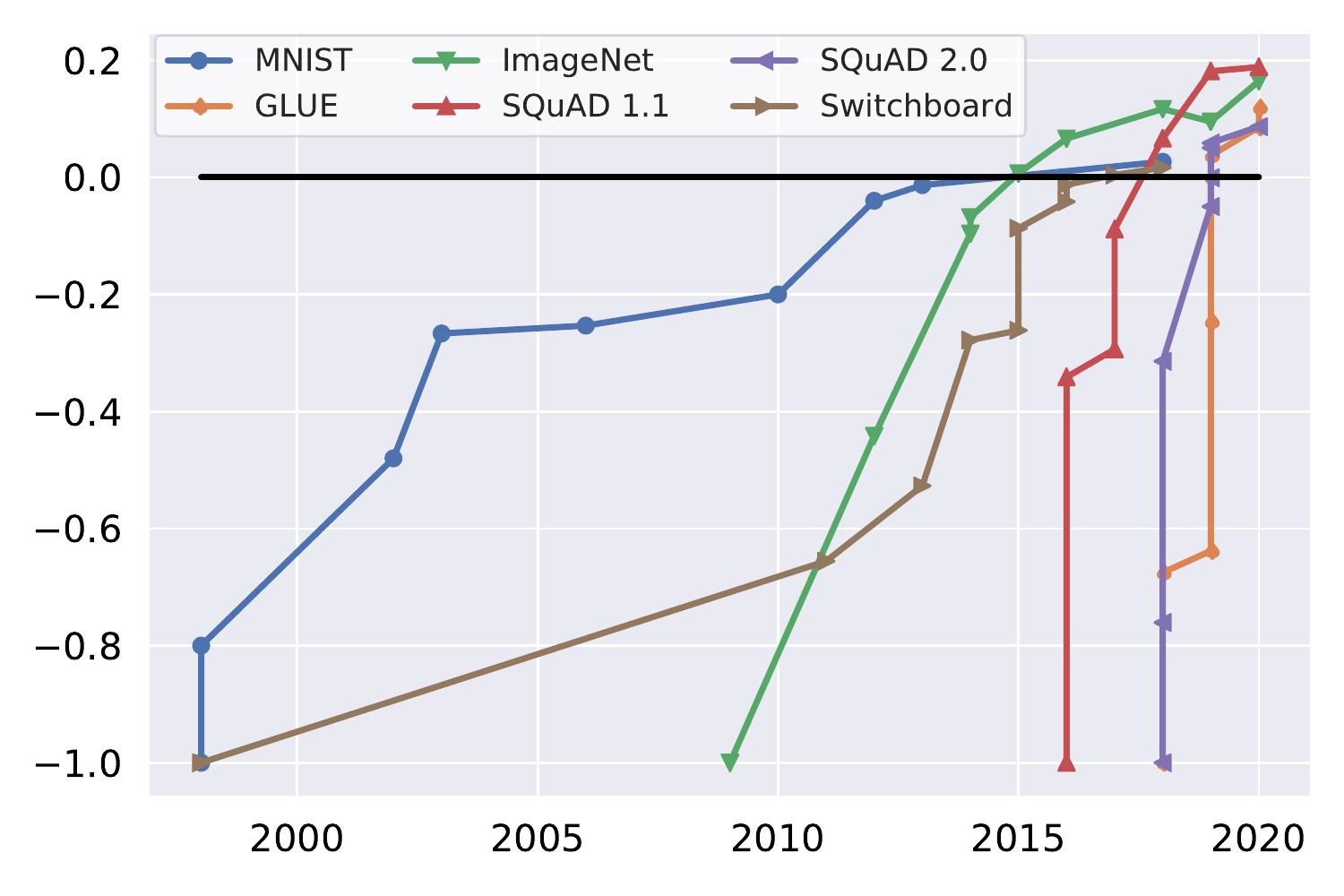}
    \caption{Benchmark saturation over time for popular benchmarks, normalized with initial performance at minus one and human performance at zero.}
    \label{fig:saturation}
\end{figure}

While it used to take decades for machine learning models to surpass estimates of human performance on benchmark tasks, that milestone is now routinely reached within just a few years for newer datasets (see Figure~\ref{fig:saturation}). As with the rest of AI, NLP has advanced rapidly thanks to improvements in computational power, as well as algorithmic breakthroughs, ranging from attention mechanisms~\cite{bahdanau2014neural, luong-etal-2015-effective}, to Transformers~\cite{vaswani2017attention}, to pre-trained language models~\cite{howard-ruder-2018-universal, devlin2019bert, liu2019roberta, radford2019gpt2, brown2020gpt3}. Equally important has been the rise of benchmarks that support the development of ambitious new data-driven models and that encourage apples-to-apples model comparisons. Benchmarks provide a north star goal for researchers, and are part of the reason we can confidently say we have made great strides in our field.
%
% In addition to compute, data, and algorithms, the fourth prong of the AI revolution has been measuring progress, or evaluating models. In particular, benchmark datasets such as MNIST, ImageNet, and GLUE have played a crucial role in driving progress.
% %
% Benchmarks, as benchmark datasets are often simply called, constitute the standard way by which models can be measured. They encourage apples-to-apples model comparisons, provide a north star goal for researchers to focus on, and are part of the reason we can confidently say that we have made great strides in our field.
%To paraphrase liberally, without ImageNet, there would be no AlexNet.
%\footnote{\url{https://qz.com/1034972/}} % - if we feel this claim needs evidence

%DK: Taking this out.. might be better suited for BG?
%Back when data was scarce and compute (for hyperparameter tuning) in short supply, k-fold cross-validation was \textit{de rigueur}. With the advent of crowd-sourcing, collecting very large-scale language datasets became feasible. The currently dominant paradigm for dataset construction and model evaluation is hold-out cross-validation, where we obtain a dataset and split it three-way into i.i.d. training, validation and test sets. The validation set is used for model selection, while the test set measures generalization.

In light of these developments, 
%
%With the field advancing as rapidly as it is, as measured by the rapid saturation of many standard static benchmark test sets, 
%DK: Pontus, I acknowledge your feedback on this para, but I've decided to keep it like this for now unless you really feel super strongly ;) I think we back up these claims in different sections of the paper and we can be a little opinionated to get the point across
one might be forgiven for thinking that NLP has created models with human-like language capabilities. Practitioners know that, despite our progress, we are actually far from this goal. Models that achieve super-human performance on benchmark tasks (according to the narrow criteria used to define human performance) nonetheless fail on simple challenge examples and falter in real-world scenarios. A substantial part of the problem is that our benchmark tasks are not adequate proxies for the sophisticated and wide-ranging capabilities we are targeting: they contain inadvertent and unwanted statistical and social biases that make them artificially easy and misaligned with our true goals.

%In fact, static benchmarks suffer from well-known issues that make it clear we have yet to achieve genuine humanlike NLU (see Section~\ref{sec:background}): 
% they saturate, do not necessarily measure what we care about, and contain inadvertent and unwanted statistical and social biases. Moreover, models that achieve super-human performance on benchmark tasks (according to the narrow criteria used to define human performance) nonetheless fail on simple challenge examples and falter in real-world scenarios, pointing to weaknesses in how we create datasets and benchmark models.

% DK: People might think this is too hype-y.. let's agree to disagree ;)
We believe the time is ripe to radically rethink benchmarking. In this paper, which both takes a position and seeks to offer a partial solution, we introduce Dynabench, an open-source, web-based research platform for dynamic data collection and model benchmarking. The guiding hypothesis behind Dynabench is that we can make even faster progress if we evaluate models and collect data dynamically, with humans and models in the loop, rather than the traditional static way.

%We propose a scientific experiment: Can we make even faster progress if we evaluate models and collect data dynamically, with humans and models in the loop, rather than in the traditional static way?

Concretely, Dynabench hosts tasks for which we dynamically collect data against state-of-the-art models in the loop, over multiple rounds. The stronger the models are and the fewer weaknesses they have, the lower their error rate will be when interacting with humans, giving us a concrete metric---i.e., how well do AI systems perform when interacting with humans? This reveals the shortcomings of state-of-the-art models, and it yields valuable training and assessment data which the community can use to develop even stronger models.

In this paper, we first document the background that led us to propose this platform. We then describe the platform in technical detail, report on findings for four initial tasks, and address possible objections. We finish with a discussion of future plans and next steps. %Dynabench is a scientific experiment that will allow us to dynamically collect high-quality data and to better assess model performance. (AW: this repeats what we say a few lines above)
%DK: this needs a better punchline

\section{Background}\label{sec:background}

Progress in NLP has traditionally been measured through a selection of %high-quality, 
task-level datasets that gradually became accepted benchmarks \citep{marcus-etal-1993-building,pradhan-etal-2012-conll}. Recent well-known examples include the Stanford Sentiment Treebank~\citep{socher-etal-2013-recursive}, SQuAD~\cite{rajpurkar2016squad, rajpurkar-etal-2018-know}, SNLI \citep{bowman-etal-2015-large}, and MultiNLI~\cite{williams-etal-2018-broad}. More recently, multi-task benchmarks such as SentEval~\cite{conneau-kiela-2018-senteval}, DecaNLP~\cite{McCann2018decaNLP}, GLUE~\cite{wang-etal-2018-glue}, and SuperGLUE~\cite{wang2019superglue} were proposed with the aim of measuring general progress across several tasks.
%
%When will we know that we have achieved human-level language capabilities? A natural direction would be to examine how well humans do on these tasks. Implicitly, the assumption would then be that if we can match or surpass human performance, as measured by these static benchmark test sets, we will know that we have achieved human-level capabilities.
%
When the GLUE dataset was introduced, ``solving GLUE'' was deemed 
``beyond the capability of current transfer learning methods''~\cite{wang-etal-2018-glue}. However, GLUE saturated within a year and its successor, SuperGLUE, already has models rather than humans at the top of its leaderboard.
%is not far off.
%
%Since nobody believes that we have achieved human-level natural language capabilities, we clearly have a measurement problem.
These are remarkable achievements, but there is an extensive body of evidence indicating that these models do not in fact have the human-level natural language capabilities one might be lead to believe.

\subsection{Challenge Sets and Adversarial Settings}
% papers that show that models are not robust/dont generalize
% incorporate somewhere: Inoculation by Fine-Tuning: A Method for Analyzing Challenge Datasets Liu et al

Whether our models have learned to solve tasks in robust and generalizable ways has been a topic of much recent interest. Challenging test sets have shown that many state-of-the-art NLP models struggle with compositionality~\cite{Nie_Wang_Bansal_2019, kim-linzen-2020-cogs,yu-ettinger-2020-assessing, white-etal-2020-universal}, and find it difficult to pass the myriad stress tests for social \citep{rudinger-etal-2018-gender, may-etal-2019-measuring, nangia-etal-2020-crows} and/or linguistic competencies~\cite{geiger2018stress, naik-etal-2018-stress, glockner-etal-2018-breaking, white-etal-2018-lexicosyntactic, warstadt-etal-2019-investigating, gauthier-etal-2020-syntaxgym, hossain-etal-2020-analysis, jeretic-etal-2020-natural, lewis2020question, saha-etal-2020-conjnli, schuster-etal-2020-harnessing, sugawara2020assessing, warstadt-etal-2020-blimp-benchmark}. Yet, challenge sets may suffer from performance instability \citep{liu-etal-2019-inoculation, rozen-etal-2019-diversify, zhou-etal-2020-curse} and often lack sufficient statistical power \citep{card-etal-2020-little}, suggesting that, although they may be valuable assessment tools, they are not sufficient for ensuring that our models have achieved the learning targets we set for them. 

%Another promising angle on model robustness measures vulnerability to various adversarial settings or attacks; m
Models are susceptible to adversarial attacks, and despite impressive task-level performance, state-of-the-art systems still struggle to learn robust representations of linguistic knowledge~\cite{ettinger-etal-2017-towards}, as also shown by work analyzing model diagnostics~\cite{ettinger-2020-bert, ribeiro-etal-2020-beyond}.
For example, question answering models can be fooled by simply adding a relevant sentence to the passage~\citep{jia2017adversarial}.

%, or encouraged to predict the original answer with even higher confidence for perturbed examples %perturbed to become unanswerable through entity swapping~\citep{welbl-etal-2020-undersensitivity}. %DK: can put this back in, but I don't understand this sentence
%
Text classification models have been shown to be sensitive to single input character change~\cite{ebrahimi2018hotflip} and first-order logic inconsistencies~\cite{minervini-riedel-2018-adversarially}.
Similarly, machine translation systems have been found susceptible to character-level perturbations~\cite{ebrahimi-etal-2018-adversarial} and synthetic and natural noise~\cite{belinkov2018synthetic, khayrallah-koehn-2018-impact}.
Natural language inference models can be fooled by simple syntactic heuristics or hypothesis-only biases~\cite{gururangan-etal-2018-annotation,poliak-etal-2018-hypothesis,tsuchiya-2018-performance,belinkov-etal-2019-dont,mccoy-etal-2019-right}.
Dialogue models may ignore perturbations of dialogue history \cite{sankar-etal-2019-neural}.
More generally, \citet{wallace-etal-2019-universal} find universal adversarial perturbations forcing targeted model errors across a range of tasks.
%
%In many of these cases, model blind spots can be at least partially overcome through adversarial training, which is encouraging.
%
%DK: commenting this out for now, but we need DROP and HellaSwag citations to go somewhere:
%Considerable effort has gone into designing more challenging evaluation settings that push contemporary models beyond their capabilities~\cite{rajpurkar-etal-2018-know, wang2019superglue, dua-etal-2019-drop, zellers-etal-2019-hellaswag}. 
Recent work has also focused on evaluating model diagnostics through counterfactual augmentation~\cite{kaushik2019learning}, decision boundary analysis~\cite{gardner-etal-2020-evaluating, swayamdipta-etal-2020-dataset}, and behavioural testing~\cite{ribeiro-etal-2020-beyond}. 

\subsection{Adversarial Training and Testing}

Research progress has traditionally been driven by a cyclical process of resource collection and architectural improvements. 
Similar to Dynabench, recent work seeks to embrace this phenomenon, addressing many of the previously mentioned issues through an iterative human-and-model-in-the-loop annotation process~\cite{yang2017mastering,dinan-etal-2019-build,chen-etal-2019-codah,bartolo2020beat,nie-etal-2020-adversarial}, to find ``unknown unknowns'' \cite{attenberg2015beat} or in a never-ending or life-long learning setting~\cite{silver2013lifelong,mitchell2018never}. The Adversarial NLI (ANLI) dataset~\cite{nie-etal-2020-adversarial}, for example, was collected with an adversarial setting over multiple rounds to yield ``a `moving post' dynamic target for NLU systems, rather than a static benchmark that will eventually saturate''. In its few-shot learning mode, GPT-3 barely shows ``signs of life''~\cite{brown2020gpt3} (i.e., it is barely above random) on ANLI, which is evidence that we are still far away from human performance on that task.

\subsection{Other Related Work}

While crowdsourcing has been a boon for large-scale NLP dataset creation \citep{snow-etal-2008-cheap, munro-etal-2010-crowdsourcing}, we ultimately want NLP systems to handle ``natural'' data~\cite{kwiatkowski-etal-2019-natural} and be ``ecologically valid''~\cite{devries2020towardsecologically}. \newcite{ethayarajh-jurafsky-2020-utility} analyze the distinction between what leaderboards incentivize and ``what is useful in practice'' through the lens of microeconomics. A natural setting for exploring these ideas might be dialogue~\cite{hancock-etal-2019-learning,shuster-etal-2020-dialogue}. Other works have pointed out misalignments between maximum-likelihood training on i.i.d.\ train/test splits and human language~\cite{linzen-2020-accelerate,stiennon2020learning}.

% recently made explicit several problems with the i.i.d. training/testing paradigm, especially with relation to pretraining. The misalignment between  and with human preferences was recently also pointed out by \cite{stiennon2020learning}, who proposed learning from human feedback as a useful mitigation strategy.
% TODO:
%[something about how we always want to measure performance with humans anyway but it's often too expensive -- e.g. see the many papers on correlating BLEU with human judgment]
% ideally we would always evaluate everything with humans in the loop

%
% This is left over from Max's bg section, can probably make more concise:
%This work seeks to bring many of the above ideas together, emphasizing the importance of high-quality large-scale training and evaluation data in a multi-task setting, governed by a paradigm designed to push the limits of contemporary models and provide insight into their learning capabilities, while allowing rapid iteration --- ensuring that individual task benchmarks are only saturated when the task, at least within the constraints of its definition, is truly ``solved''.
%
%On a benchmark level, we aim to create a flexible task selection process such that ``solved'' tasks are adapted, expanded and/or replaced as needed in an attempt to provide an accurate snapshot of NLP progress at any point in time.
%

We think there is widespread agreement that something has to change about our standard evaluation paradigm and that we need to explore alternatives. The persistent misalignment between benchmark performance and performance on challenge and adversarial test sets reveals that standard evaluation paradigms overstate the ability of our models to perform the tasks we have set for them. Dynabench offers one path forward from here, by allowing researchers to combine model development with the stress-testing that needs to be done to achieve true robustness and generalization.

\begin{figure*}[t]
    \centering
    \includegraphics[width=\textwidth]{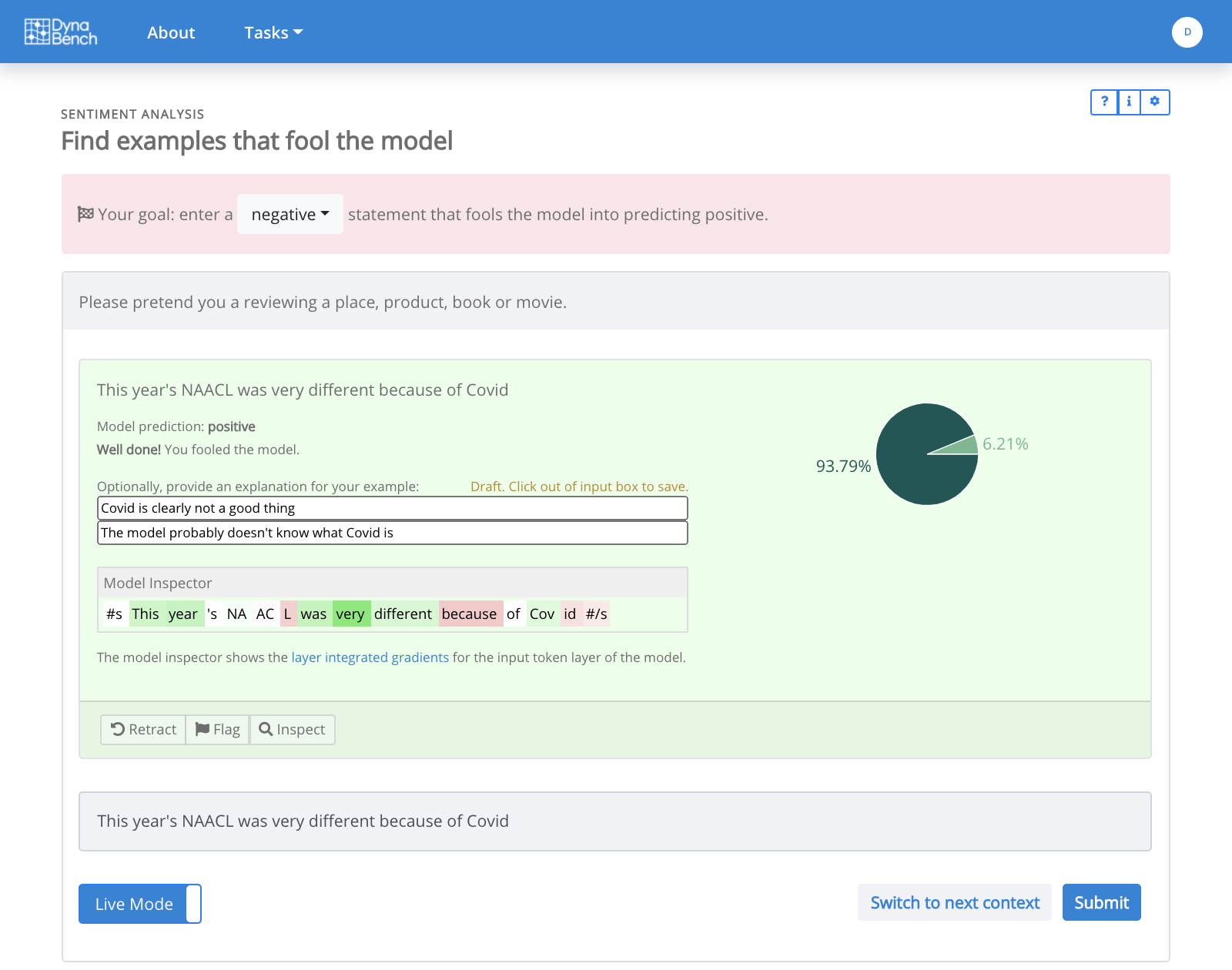}
    \caption{The Dynabench example creation interface for sentiment analysis with illustrative example.}
    \label{fig:example1}
\end{figure*}

\section{Dynabench}

Dynabench is a platform that encompasses different \emph{tasks}. Data for each task is collected over multiple \emph{rounds}, each starting from the current state of the art. In every round, we have one or more \emph{target models} ``in the loop.'' These models interact with humans, be they expert linguists or crowdworkers, who are in a position to identify models' shortcomings by providing \emph{examples} for an optional \emph{context}. Examples that models get wrong, or struggle with, can be validated by other humans to ensure their correctness. The data collected through this process can be used to evaluate state-of-the-art models, and to train even stronger ones, hopefully creating a virtuous cycle that helps drive progress in the field.  Figure~\ref{fig:example1} provides a sense of what the example creation interface looks like.

As a large-scale collaborative effort, the platform is meant to be a platform technology for human-and-model-in-the-loop evaluation that belongs to the entire community. In the current iteration, the platform is set up for dynamic \emph{adversarial} data collection, where humans can attempt to find model-fooling examples. This design choice is due to the fact that the \emph{average} case, as measured by maximum likelihood training on i.i.d.\ datasets, is much less interesting than the \emph{worst} (i.e., adversarial) case, which is what we want our systems to be able to handle if they are put in critical systems where they interact with humans in real-world settings.

However, Dynabench is not limited to the adversarial setting, and one can imagine scenarios where humans are rewarded not for fooling a model or ensemble of models, but for finding examples that models, even if they are right, are very uncertain about, perhaps in an active learning setting. Similarly, the paradigm is perfectly compatible with collaborative settings that utilize human feedback, or even negotiation. The crucial aspect of this proposal is the fact that models and humans interact live ``in the loop'' for evaluation and data collection.

One of the aims of this platform is to put expert linguists center stage. Creating model-fooling examples is not as easy as it used to be, and finding interesting examples is rapidly becoming a less trivial task. In ANLI, the verified model error rate for crowd workers in the later rounds went below 1-in-10 \citep{nie-etal-2020-adversarial}, 
while in ``Beat the AI'', human performance decreased while time per valid adversarial example went up with stronger models in the loop~\citep{bartolo2020beat}.
For expert linguists, we expect the model error to be much higher, but if the platform actually lives up to its virtuous cycle promise, that error rate will go down quickly. Thus, we predict that linguists with expertise in exploring the decision boundaries of machine learning models will become essential. 
% To turn the Fred Jelinek quote on its head, every time we hire a linguist, we expect the performance of our models to go up.
%
% CP: The direct quotation seemed a bit misleading or potentially confusing, so I reworded it.
%``every time I hire a linguist, the performance of my NLP system goes up.''

%
While we are primarily motivated by evaluating progress, both ANLI and ``Beat the AI'' show that models can overcome some of their existing blind spots through adversarial training.
They also find that best model performance is still quite far from that of humans, suggesting that while the collected data appears to lie closer to the model decision boundaries, there still exist adversarial examples beyond the remit of current model capabilities.

\subsection{Features and Implementation Details}
Dynabench offers low-latency, real-time feedback on the behavior of state-of-the-art NLP models. The technology stack is based on PyTorch \cite{Pytorch:2019neurips}, with models served via TorchServe.\footnote{\url{https://pytorch.org/serve}}
The platform not only displays prediction probabilities, but through an ``inspect model'' functionality, allows the user to examine the token-level layer integrated gradients \citep{sundararajan-etal-2017-axiomatic}, obtained via the Captum interpretability library.\footnote{\url{https://captum.ai/}}

For each example, we allow the user to \emph{explain} what the correct label is, as well as why they think it fooled a model if the model got it wrong; or why the model might have been fooled if it wasn't. All collected model-fooling (or, depending on the task, even non-model-fooling) examples are verified by other humans to ensure their validity.

Task owners can collect examples through the web interface, by engaging with the community, or through Mephisto,\footnote{\url{https://github.com/facebookresearch/Mephisto}} which makes it easy to connect, e.g., Mechanical Turk workers to the exact same backend. All collected data will be open sourced, in an anonymized fashion.

In its current mode, Dynabench could be described as a fairly conservative departure from the status quo. It is being used to develop datasets that support the same metrics that drive existing benchmarks. The crucial change is that the datasets are now dynamically created, allowing for more kinds of evaluation---e.g., tracking progress through rounds and across different conditions.% As Dynabench evolves, we expect to see a greater variety of evaluation techniques.

\subsection{Initial Tasks}
We have selected four official tasks as a starting point, which we believe represent an appropriate cross-section of the field at this point in time.
Natural Language Inference (NLI) and Question Answering (QA) are canonical tasks in the field. Sentiment analysis is a task that some consider ``solved''~(and is definitely treated as such, with all kinds of ethically problematic repercussions), which we show is not the case. Hate speech is very important as it can inflict harm on people, yet classifying it remains challenging for NLP.

% !!! TASK OWNERS - See below !!!
% What to put here? What the task is, what the current round does, what is special about this task or what the team is focusing on?
% IMPORTANT NOTE FROM MOHIT: Hi everyone, please use they/their etc. instead of we/our everywhere for our previous ANLI/BeatTheAI etc. works to avoid anonymity violation rules/desk-reject, thanks :)

\paragraph{Natural language inference.}
%
%TBD - Yixin / Mohit / Grusha / Adina
Built upon the semantic foundation of natural logic \cite[i.a.]{sanchez1991studies}\ and hailing back much further \citep{vanBenthem08NATLOG},
%at least to Richard Montague~\citep{montague1970universal}, 
NLI is one of the quintessential natural language understanding tasks. NLI, also known as `recognizing textual entailment'~\citep{dagan2006}, is often formulated as a 3-way classification problem where the input is a context sentence paired with a hypothesis, and the output is a label (entailment, contradiction, or neutral) indicating the relation between the pair.

% I think it should sound more like, ANLI is prior work that used a similar interface to collect 3 rounds of data. We start with their data + models and collect an additional round in Dynabench, with modifications x, y, and z.
%The first three rounds of NLI data were collected prior to the launch of Dynabench as a part of ANLI.
% With permission from the authors, w
We build on the ANLI dataset \citep{nie-etal-2020-adversarial} and its three rounds to seed the Dynabench NLI task. During the ANLI data collection process, the annotators were presented with a context (extracted from a pre-selected corpus) and a desired target label, and asked to provide a hypothesis that fools the target model adversary into misclassifying the example. If the target model is fooled, the annotator was invited to speculate about why, or motivate why their example was right. The target model of the first round (R1) was a single BERT-Large model fine-tuned on SNLI and MNLI, while the target model of the second and third rounds (R2, R3) was an ensemble of RoBERTa-Large models fine-tuned on SNLI, MNLI, FEVER~\citep{thorne-etal-2018-fever} recast as NLI, and all of the ANLI data collected prior to the corresponding round. The contexts for Round~1 and Round~2 were Wikipedia passages curated in~\citet{yang-etal-2018-hotpotqa} and the contexts for Round 3 were from various domains. Results indicate that state-of-the-art models~(which can obtain 90\%+ accuracy on SNLI and MNLI) cannot exceed 50\% accuracy on rounds 2 and 3.

With the launch of Dynabench, we have started collection of a fourth round, which has several innovations: not only do we select candidate contexts from a more diverse set of Wikipedia featured articles but we also use an ensemble of two different models with different architectures as target adversaries to increase diversity and robustness. Moreover, the ensemble of adversaries will help mitigate issues with creating a dataset whose distribution is too closely aligned to a particular target model or architecture. Additionally, we are collecting two types of natural language explanations: why an example is correct and why a target model might be wrong. We hope that disentangling this information will yield an additional layer of interpretability and yield models that are as least as explainable as they are robust.

\paragraph{Question answering.}
% TBD - Max / Pontus / Sebastian / Divyansh / Robin
The QA task takes the same format as SQuAD1.1~\cite{rajpurkar2016squad}, i.e., given a context and a question, extract an answer from the context as a continuous span of text.
The first round of adversarial QA (AQA) data comes from ``Beat the AI''~\cite{bartolo2020beat}.
During annotation, crowd workers were presented with a context sourced from Wikipedia, identical to those in SQuAD1.1, and 
asked to write a question and select an answer. The annotated answer was compared to the model prediction using a word-overlap F$_1$ threshold and, if sufficiently different, considered to have fooled the model.
The target models in round 1 were BiDAF~\cite{Seo2016BidAF}, BERT-Large, and RoBERTa-Large.
%
% Results find a performance gap between humans and model baselines, particularly with stronger models-in-the-loop.
%

%
The model in the loop for the current round is RoBERTa trained on the examples from the first round combined with SQuAD1.1.
Despite the super-human performance achieved on SQuAD1.1, machine performance is still far from humans on the current leaderboard.
In the current phase, we seek to collect rich and diverse examples, focusing on improving model robustness through generative data augmentation, to provide more challenging model adversaries in this constrained task setting.
We should emphasize that we don't consider this task structure representative of the broader definition even of closed-domain QA, and are looking to expand this to include unanswerable questions~\cite{rajpurkar-etal-2018-know}, longer and more complex passages, Yes/No questions and multi-span answers~\cite{kwiatkowski-etal-2019-natural}, and numbers, dates and spans from the question~\cite{dua-etal-2019-drop} as model performance progresses.

\paragraph{Sentiment analysis.}
%
%TBD - Zen / Atticus / Chris

The sentiment analysis project is a multi-pronged effort to create a dynamic benchmark for sentiment analysis and to evaluate some of the core hypotheses behind Dynabench. \citet{potts-etal-2020-dynasent} provide an initial report and the first two rounds of this dataset.

The task is structured as a 3-way classification problem: positive, negative, and neutral. The motivation for using a simple positive/negative dichotomy is to show that there are still very challenging phenomena in this traditional sentiment space. The neutral category was added to avoid~(and helped trained models avoid) the false presupposition that every text conveys sentiment information \citep{PangLee08}.  In future iterations, we plan to consider additional dimensions of sentiment and emotional expression \citep{alm-etal-2005-emotions, neviarouskaya-etal-2010-recognition, Wiebe:Wilson:Cardie:2005,Liu+Lieberman+Selker:03a, Sudhof-etal:2014}.

In this first phase, we examined the question of how best to elicit examples from workers that are diverse, creative, and naturalistic. In the ``prompt'' condition, we provide workers with an actual sentence from an existing product or service review and ask them to edit it so that it fools the model. In the ``no prompt'' condition, workers try to write original sentences that fool the model. We find that the ``prompt'' condition is superior: workers generally make substantial edits, and the resulting sentences are more linguistically diverse than those in the ``no prompt'' condition.

In a parallel effort, we also collected and validated hard sentiment examples from existing corpora, which will enable another set of comparisons that will help us to refine the Dynabench protocols and interfaces. We plan for the dataset to continue to grow, probably mixing attested examples with those created on Dynabench with the help of prompts. With these diverse rounds, we can address a wide range of question pertaining to dataset artifacts, domain transfer, and overall robustness of sentiment analysis systems. 

\paragraph{Hate speech detection.}
The hate speech task classifies whether a statement expresses hate against a protected characteristic or not. Detecting hate is notoriously difficult given the important role played by context and speaker \cite{LeaderMaynard2016} and the variety of ways in which hate can be expressed \cite{Waseem2017}. Few high-quality, varied and large training datasets are available for training hate detection systems~\cite{Vidgen2020, Poletto2020, vidgen-etal-2019-challenges}. %Humans often do not agree on what hate is \cite{Salminen2019}. %Many 'state of the art' classification systems perform poorly in out-of-domain contexts \cite{Karan2018, Salminen2020}. Neutral content is often misclassified as hate if it contains any mention of an identity, counter speech  \cite[i.e., content that challenges hate speech;][]{Mathew2019}, incivil language, negative sentiment (not directed at an identity) or interpersonal attacks. Hate is often mistaken for neutral in cases of subtle hate, abuse against a less common identity, or coded, misspelled or unusual text \cite{Fortuna2018b, vidgen_challenges_2019, schmidtSurveyHateSpeech2017a}.

We organised four rounds of data collection and model training, with preliminary results reported in \newcite{vidgen2020learning}. In each round, annotators are tasked with entering content that tricks the model into giving an incorrect classification. The content is created by the annotators and as such is synthetic in nature.
At the end of each round the model is retrained and the process is repeated. For the first round, we trained a RoBERTa model on ~470,000 hateful and abusive statements\footnote{Derived from \url{https://hatespeechdata.com}, in anonymized form.}. 
For subsequent rounds the model was trained on the original data plus content from the prior rounds. Due to the complexity of online hate, we hired and trained analysts rather than paying for crowd-sourced annotations. Each analyst was given training, support, and feedback throughout their work.

In all rounds annotators provided a label for whether content is hateful or not. In rounds 2, 3 and 4, they also gave labels for the target (i.e., which group has been attacked) and type of statement (e.g., derogatory remarks, dehumanization, or threatening language). These granular labels help to investigate model errors and improve performance, as well as directing the identification of new data for future entry. For approximately half of entries in rounds 2, 3 and 4, annotators created ``perturbations'' where the text is minimally adjusted so as to flip the label~\cite{gardner-etal-2020-evaluating,kaushik2019learning}. This helps to identify decision boundaries within the model, and minimizes the risk of overfitting given the small pool of annotators.

Over the four rounds, content becomes increasingly adversarial (shown by the fact that target models have lower performance on later rounds' data) and models improve (shown by the fact that the model error rate declines and the later rounds' models have the highest accuracy on each round). We externally validate performance using the \textsc{HateCheck} suite of diagnostic tests from \newcite{rottger2020hatecheck}. We show substantial improvement over the four rounds, and our final round target model achieves 94\% on \textsc{HateCheck}, outperforming the models presented by the original authors. %This suggests that our approach can be used to create high-performing, robust and generalisable hate detection systems.

\begin{table}[t]
%\small
    \centering
    \begin{tabular}{lrrr}
        \toprule
        Task & Rounds & Examples & vMER\\
        \midrule
        NLI & 4 & 170,294 & 33.24\%\\
        QA & 2 & 36,406 & 33.74\%\\
        Sentiment & 3 & 19,975 & 35.00\%\\
        Hate speech & 4 & 41,255 & 43.90\%\\
        % Task & Round & Examples & Target & vMER\\
        % \midrule
        % NLI & 4 & 170,294$^\dagger$ & Ensemble & 44.12\%\\
        % QA & 2 & 36,406$^\ddagger$ & RoBERTa & 33.74\%\\
        % Sentiment & 1 & 5,081 & RoBERTa & 47.65\%\\
        % Hate speech & 2 & 20,819 & RoBERTa & 56.75\%\\
        \bottomrule
    \end{tabular}
    \caption{Statistics for the initial four official tasks. %$\dagger$ includes Adversarial NLI \cite{nie-etal-2020-adversarial}; $\ddagger$ includes Beat the AI \cite{bartolo2020beat}.
    %The NLI target ensemble comprises RoBERTa, ALBERT, BART, ELECTRA and XLNet; the other tasks have a RoBERTa target model.
    }
    \label{tab:results}
\end{table}

\subsection{Dynabenchmarking NLP}

Table~\ref{tab:results} shows an overview of the current situation for the four tasks. Some tasks are further along in their data collection efforts than others. As we can see, the validated model error rate (vMER; the number of human-validated model errors divided by the total number of examples---note that the error rates are not necessarily comparable across tasks, since the interfaces and in-the-loop models are not identical) is still very high across all tasks, clearly demonstrating that NLP is far from solved.
%TODO: show that as rounds progress, show what happens to the vMER  - perhaps this is best done in a comprehensive follow-up study?

\section{Caveats and Objections}

There are several obvious and valid objections one can raise. We do not have all the answers, but we can try to address some common concerns.

\paragraph{Won't this lead to unnatural distributions and distributional shift?} 

% CP: I suggested a revision just below:
%
% Yes, it probably will. Obviously, data collected in any ``unnatural'' setting for the purpose of data collection (including static data collection using crowd-workers) is not natural, and if we collect data using dynamic adversarial collection this problem might be exacerbated by annotator incentives. We do not believe that models should ever be trained \emph{only} on adversarial data, or even \emph{only} evaluated on adversarial data. Rather, next-generation models should do better on dynamic adversarial data \emph{without sacrificing performance} on static benchmarks. To use an example from NLI: it would not make sense to only train on ANLI or only evaluate on ANLI, when such high quality datasets as SNLI and MNLI are available.
% %
% For QA, \citet{bartolo2020beat} show that while training solely on adversarially-collected data was detrimental to performance on non-adversarially collected data, models are capable of simultaneously learning both distributions when trained on the combined data, retaining if not slightly improving performance on the original distribution.
% %

% %
% Note that distributional shift is a natural phenomenon that actually happens quite a lot in natural language (think e.g. of the meaning of ``rock'' and ``gay'' over time), and that this is something we are going to have to solve anyway from a machine learning perspective. We hope to inspire further research into characterizing and being robust to distributional shift.

Yes, that is a real risk. First, we acknowledge that crowdsourced texts are likely to have unnatural qualities: the setting itself is artificial from the perspective of genuine communication, and crowdworkers are not representative of the general population. Dynabench could exacerbate this, but it also has features that can help alleviate it. For instance, as we discussed earlier, the sentiment analysis project is using naturalistic prompt sentences to try to help workers create more diverse and naturalistic data.

Second, if we rely solely on dynamic \emph{adversarial} collection, then we increase the risks of creating unnatural datasets. For instance, \citet{bartolo2020beat} show that training solely on adversarially-collected data for QA was detrimental to performance on non-adversarially collected data. However, they also show that models are capable of simultaneously learning both distributions when trained on the combined data, retaining if not slightly improving performance on the original distribution (of course, this may not hold if we have many more examples of one particular kind). Ideally, we would combine adversarially collected data with non-adversarial---preferably naturally collected---data, so as to capture both the average and worst case scenarios in our evaluation.

Finally, we note that Dynabench could enable the community to explore the kinds of distributional shift that are characteristic of natural languages. Words and phrases change their meanings over time, between different domains, and even between different interlocutors. Dynabench could be a tool for studying such shifts and finding models that can succeed on such phenomena.

\paragraph{What if annotators ``overfit'' on models?}
A potential risk is cyclical ``progress,'' where improved models forget things that were relevant in earlier rounds because annotators focus too much on a particular weakness. Continual learning is an exciting research direction here: we should try to understand distributional shift better, as well as how to characterize how data shifts over time might impact learning, and how any adverse effects might be overcome. Because of how most of us have been trained, it is natural to assume that the \emph{last} round is automatically the best evaluation round, but that does not mean that it should be the only round: in fact, most likely, the best way to evaluate progress is to evaluate on \emph{all} rounds as well as \emph{any} high-quality static test set that exists, possibly with a recency-based discount factor. To make an analogy with software testing, similar to checklists~\cite{ribeiro-etal-2020-beyond}, it would be a bad idea to throw away old tests just because you've written some new ones. As long as we factor in previous rounds, Dynabench's dynamic nature offers a way out from forgetting and cyclical issues: any model biases will be fixed in the limit by annotators exploiting vulnerabilities.

Another risk is that the data distribution might be too heavily dependent on the target model in the loop. When this becomes an issue, it can be mitigated by using ensembles of many different architectures in the loop, for example the top current state-of-the-art ones, with multiple seeds.\footnote{ANLI does not show dramatically different results across models, suggesting that this is not necessarily a big problem yet, but it shows in R2 and R3 that ensembles are possible.}

% wrt naturalness:
%\cite{kwiatkowski-etal-2019-natural}

%Remember that Dynabench is a scientific experiment! A few things will help, however: we can make sure we don't shift in the direction of one particular model by using ensembles as adversarial targets, by validating collected examples with other humans and by not discarding data from older rounds. That said, it is unclear to what extent a statically collected crowdsourced datasets are more "natural".

\paragraph{How do we account for future, not-yet-in-the-loop models?}

Obviously, we can't---so this is a very valid criticism. However, we can assume that an ensemble of model architectures is a reasonable approximation, if and only if the models are not too bad at their task. This latter point is crucial: we take the stance that models by now, especially in aggregate, are probably good enough to be reasonably close enough to the decision boundaries---but it is definitely true that we have no guarantees that this is the case.
%So far, there is evidence that adversarially collected data provides performance gains irrespective of the model in the loop \cite{nie-etal-2020-adversarial,dinan-etal-2019-build,bartolo2020beat}.
%^ grusha.prasad: This point seems like it is good empirical evidence for "how do we account for not-yet-in-the-loop models"? So saying, "there is no guarantee that we can, but existing evidence suggests that collecting adversarial data with the best available models will likely be beneficial for future models. And if they are not, then the future models can be added in the loop for a new round of data collection"
%DK: I agree, let's revisit (I think Beat the AI has the clearest story here?)

\paragraph{How do we compare results if the benchmark keeps changing?} This is probably the main hurdle from a community adoption standpoint. But if we consider,  e.g., the multiple iterations of SemEval or WMT datasets over the years, we've already been handling this quite well---we accept that a model's BLEU score on WMT16 is not comparable to WMT14. That is, it is perfectly natural for benchmark datasets to evolve as the community makes progress. The only thing Dynabench does differently is that it \emph{anticipates} dataset saturation and embraces the loop so that we can make faster and more sustained progress. % without having to go in search of a new benchmark every time one saturates.

\paragraph{What about generative tasks?} For now Dynabench focuses on classification or span extraction tasks where it is relatively straightforward to establish whether a model was wrong. If instead the evaluation metric is something like ROUGE or BLEU and we are interested in generation, we need a way to discretize an answer to determine correctness, since we wouldn't have ground truth annotations; which makes determining whether a model was successfully fooled less straightforward. However, we could discretize generation by re-framing it as multiple choice with hard negatives, or simply by asking the annotator if the generation is good enough. In short, going beyond classification will require further research, but is definitely doable.

\paragraph{Do we need models in the loop for good data?} The potential usefulness of adversarial examples can be explained at least in part by the fact that having an annotation partner (so far, a model) simply provides better incentives for generating quality annotation. Having the model in the loop is obviously useful for evaluation, but it's less clear if the resultant data is necessarily also useful in general for training.
So far, there is evidence that adversarially collected data provides performance gains irrespective of the model in the loop \cite{nie-etal-2020-adversarial,dinan-etal-2019-build,bartolo2020beat}. For example, ANLI shows that replacing equal amounts of ``normally collected'' SNLI and MNLI training data with ANLI data improves model performance, especially when training size is small~\cite{nie-etal-2020-adversarial}, suggesting higher data efficiency. However, it has also been found that model-in-the-loop counterfactually-augmented training data does not necessarily lead to better generalization~\cite{huang2020counterfactually}. Given the distributional shift induced by adversarial settings, it would probably be wisest to combine adversarially collected data with non-adversarial data during training (ANLI takes this approach), and to also test models in both scenarios.
To get the most useful training and testing data, it seems the focus should be on collecting adversarial data with the best available model(s), preferably with a wide range of expertise, as that will likely be beneficial to future models also. That said, we expect this to be both task and model dependent. Much more research is required, and we encourage the community to explore these topics.

% CP suggested revision below. It would be great to get more numbers from other projects!
%
% \paragraph{This is too expensive. How will it scale?} Model builders are incentivized to use Dynabench to showcase their latest model advances. We hope that similarly, elite model breakers will appreciate the honor that comes with being high on the user leaderboard for breaking models. In addition, we are working on making the tool available for education, as well as gamifying the interface to make it (even) more fun to try to fool models. But ultimately, yes: this will always need some form of funding, which given the progress it promises should not be too difficult to secure.

% CP revision:
%
\paragraph{Is it expensive?}
Dynamic benchmarking is indeed expensive, but it is worth putting the numbers in context, as all data collection efforts are expensive when done at the scale of our current benchmark tasks. For instance, SNLI has 20K examples that were separately validated, and each one of these examples cost approximately \$0.50 to obtain and validate (personal communication with SNLI authors). Similarly, the 40K validated examples in MultiNLI cost \$0.64 each (p.c., MultiNLI authors). By comparison, the average cost of creation and validation for ANLI examples is closer to \$1.00 (p.c., ANLI authors). This is a substantial increase at scale. However, dynamic adversarial datasets may also last longer as benchmarks. If true, then the increased costs could turn out to be a bargain.

We should acknowledge, though, that dynamic benchmarks will tend to be more expensive than regular benchmarks for comparable tasks, because not every annotation attempt will be model-fooling and validation is required. Such expenses are likely to increase through successive rounds, as the models become more robust to workers' adversarial attacks. The research bet is that each example obtained this way is actually worth more to the community and thus worth the expense.

In addition, we hope that language enthusiasts and other non-crowdworker model breakers will appreciate the honor that comes with being high up on the user leaderboard for breaking models. We are working on making the tool useful for education, as well as gamifying the interface to make it (even) more fun to try to fool models, as a ``game with a purpose'' \cite{vonahn2008designing}, for example through the ability to earn badges.

\section{Conclusion and Outlook}

We introduced Dynabench, a research platform for dynamic benchmarking. Dynabench opens up exciting new research directions, such as investigating the effects of ensembles in the loop, distributional shift characterisation, exploring annotator efficiency, investigating the effects of annotator expertise, and improving model robustness to targeted adversarial attacks in an interactive setting. It also facilitates further study in dynamic data collection, and more general cross-task analyses of human-and-machine interaction. The current iteration of the platform is only just the beginning of a longer journey. In the immediate future, we aim to achieve the following goals:

\paragraph{Anyone can run a task.} Having created a tool that allows for human-in-the-loop model evaluation and data collection, we aim to make it possible for anyone to run their own task. To get started, only three things are needed: a target model, a (set of) context(s), and a pool of annotators.

\paragraph{Multilinguality and multimodality.} As of now, Dynabench is text-only and focuses on English, but we hope to change that soon.

\paragraph{Live model evaluation.} Model evaluation should not be about one single number on some test set. If models are uploaded through a standard interface, they can be scored automatically along many dimensions. We would be able to capture not only accuracy, for example, but also usage of computational resources, inference time, fairness, and many other relevant dimensions. This will in turn enable dynamic leaderboards, for example based on utility \cite{ethayarajh-jurafsky-2020-utility}. This would also allow for backward-compatible comparisons, not having to worry about the benchmark changing, and automatically putting new state of the art models in the loop, addressing some of the main objections.

One can easily imagine a future where, in order to fulfill reproducibility requirements, authors do not only link to their open source codebase but also to their model inference point so others can ``talk with'' their model. This will help drive progress, as it will allow others to examine models' capabilities and identify failures to address with newer even better models. If we cannot always democratize the \emph{training} of state-of-the-art AI models, at the very least we can democratize their \emph{evaluation}.

\section*{Acknowledgements}
% Other people DK talked to (once): Sam Bowman, Emily Bender
% Should we add?
We would like to thank Jason Weston, Emily Dinan and Kyunghyun Cho for their input on this project, and Sonia Kris for her support. 
ZW has been supported in part by the Canada 150 Research Chair program and the UK-Canada AI Artificial Intelligence Initiative.
YN and MB have been supported in part by DARPA MCS N66001-19-2-4031, DARPA YFA17-D17AP00022, and ONR N00014-18-1-2871.
CP has been supported in part by grants from Facebook, Google, and by Stanford's Institute for Human-Centered AI.

\bibliography{anthology,references}

\begin{thebibliography}{104}
\expandafter\ifx\csname natexlab\endcsname\relax\def\natexlab#1{#1}\fi

\bibitem[{Alm et~al.(2005)Alm, Roth, and Sproat}]{alm-etal-2005-emotions}
Cecilia~Ovesdotter Alm, Dan Roth, and Richard Sproat. 2005.
\newblock \href {https://www.aclweb.org/anthology/H05-1073} {Emotions from
  text: Machine learning for text-based emotion prediction}.
\newblock In \emph{Proceedings of Human Language Technology Conference and
  Conference on Empirical Methods in Natural Language Processing}, pages
  579--586, Vancouver, British Columbia, Canada. Association for Computational
  Linguistics.

\bibitem[{Attenberg et~al.(2015)Attenberg, Ipeirotis, and
  Provost}]{attenberg2015beat}
Joshua Attenberg, Panos Ipeirotis, and Foster Provost. 2015.
\newblock \href {https://doi.org/10.1145/2700832} {Beat the machine:
  Challenging humans to find a predictive model's “unknown unknowns”}.
\newblock \emph{J. Data and Information Quality}, 6(1).

\bibitem[{Bahdanau et~al.(2014)Bahdanau, Cho, and Bengio}]{bahdanau2014neural}
Dzmitry Bahdanau, Kyunghyun Cho, and Yoshua Bengio. 2014.
\newblock \href {https://arxiv.org/abs/1409.0473} {Neural machine translation
  by jointly learning to align and translate}.
\newblock \emph{arXiv preprint arXiv:1409.0473}.

\bibitem[{Bartolo et~al.(2020)Bartolo, Roberts, Welbl, Riedel, and
  Stenetorp}]{bartolo2020beat}
Max Bartolo, Alastair Roberts, Johannes Welbl, Sebastian Riedel, and Pontus
  Stenetorp. 2020.
\newblock \href {https://doi.org/10.1162/tacl\_a\_00338} {Beat the ai:
  Investigating adversarial human annotation for reading comprehension}.
\newblock \emph{Transactions of the Association for Computational Linguistics},
  8:662--678.

\bibitem[{Belinkov and Bisk(2018)}]{belinkov2018synthetic}
Yonatan Belinkov and Yonatan Bisk. 2018.
\newblock \href {https://openreview.net/forum?id=BJ8vJebC-} {Synthetic and
  natural noise both break neural machine translation}.
\newblock In \emph{International Conference on Learning Representations}.

\bibitem[{Belinkov et~al.(2019)Belinkov, Poliak, Shieber, Van~Durme, and
  Rush}]{belinkov-etal-2019-dont}
Yonatan Belinkov, Adam Poliak, Stuart Shieber, Benjamin Van~Durme, and
  Alexander Rush. 2019.
\newblock \href {https://doi.org/10.18653/v1/P19-1084} {Don{'}t take the
  premise for granted: Mitigating artifacts in natural language inference}.
\newblock In \emph{Proceedings of the 57th Annual Meeting of the Association
  for Computational Linguistics}, pages 877--891, Florence, Italy. Association
  for Computational Linguistics.

\bibitem[{Bowman et~al.(2015)Bowman, Angeli, Potts, and
  Manning}]{bowman-etal-2015-large}
Samuel~R. Bowman, Gabor Angeli, Christopher Potts, and Christopher~D. Manning.
  2015.
\newblock \href {https://doi.org/10.18653/v1/D15-1075} {A large annotated
  corpus for learning natural language inference}.
\newblock In \emph{Proceedings of the 2015 Conference on Empirical Methods in
  Natural Language Processing}, pages 632--642, Lisbon, Portugal. Association
  for Computational Linguistics.

\bibitem[{Brown et~al.(2020)Brown, Mann, Ryder, Subbiah, Kaplan, Dhariwal,
  Neelakantan, Shyam, Sastry, Askell et~al.}]{brown2020gpt3}
Tom~B Brown, Benjamin Mann, Nick Ryder, Melanie Subbiah, Jared Kaplan, Prafulla
  Dhariwal, Arvind Neelakantan, Pranav Shyam, Girish Sastry, Amanda Askell,
  et~al. 2020.
\newblock \href {https://arxiv.org/abs/2005.14165} {Language models are
  few-shot learners}.
\newblock \emph{arXiv preprint arXiv:2005.14165}.

\bibitem[{Card et~al.(2020)Card, Henderson, Khandelwal, Jia, Mahowald, and
  Jurafsky}]{card-etal-2020-little}
Dallas Card, Peter Henderson, Urvashi Khandelwal, Robin Jia, Kyle Mahowald, and
  Dan Jurafsky. 2020.
\newblock \href {https://www.aclweb.org/anthology/2020.emnlp-main.745} {With
  little power comes great responsibility}.
\newblock In \emph{Proceedings of the 2020 Conference on Empirical Methods in
  Natural Language Processing (EMNLP)}, pages 9263--9274, Online. Association
  for Computational Linguistics.

\bibitem[{Chen et~al.(2019)Chen, D{'}Arcy, Liu, Fernandez, and
  Downey}]{chen-etal-2019-codah}
Michael Chen, Mike D{'}Arcy, Alisa Liu, Jared Fernandez, and Doug Downey. 2019.
\newblock \href {https://doi.org/10.18653/v1/W19-2008} {{CODAH}: An
  adversarially-authored question answering dataset for common sense}.
\newblock In \emph{Proceedings of the 3rd Workshop on Evaluating Vector Space
  Representations for {NLP}}, pages 63--69, Minneapolis, USA. Association for
  Computational Linguistics.

\bibitem[{Conneau and Kiela(2018)}]{conneau-kiela-2018-senteval}
Alexis Conneau and Douwe Kiela. 2018.
\newblock \href {https://www.aclweb.org/anthology/L18-1269} {{S}ent{E}val: An
  evaluation toolkit for universal sentence representations}.
\newblock In \emph{Proceedings of the Eleventh International Conference on
  Language Resources and Evaluation ({LREC} 2018)}, Miyazaki, Japan. European
  Language Resources Association (ELRA).

\bibitem[{Dagan et~al.(2006)Dagan, Glickman, and Magnini}]{dagan2006}
Ido Dagan, Oren Glickman, and Bernardo Magnini. 2006.
\newblock \href {https://link.springer.com/chapter/10.1007/11736790_9} {The
  {PASCAL} recognising textual entailment challenge}.
\newblock In \emph{Machine learning challenges. evaluating predictive
  uncertainty, visual object classification, and recognising tectual
  entailment}, pages 177--190. Springer.

\bibitem[{de~Vries et~al.(2020)de~Vries, Bahdanau, and
  Manning}]{devries2020towardsecologically}
Harm de~Vries, Dzmitry Bahdanau, and Christopher Manning. 2020.
\newblock \href {https://arxiv.org/abs/2007.14435} {Towards ecologically valid
  research on language user interfaces}.
\newblock \emph{arXiv preprint arXiv:2007.14435}.

\bibitem[{Devlin et~al.(2019)Devlin, Chang, Lee, and
  Toutanova}]{devlin2019bert}
Jacob Devlin, Ming-Wei Chang, Kenton Lee, and Kristina Toutanova. 2019.
\newblock \href {https://doi.org/10.18653/v1/N19-1423} {{BERT}: Pre-training of
  deep bidirectional transformers for language understanding}.
\newblock In \emph{Proceedings of the 2019 Conference of the North {A}merican
  Chapter of the Association for Computational Linguistics: Human Language
  Technologies, Volume 1 (Long and Short Papers)}, pages 4171--4186,
  Minneapolis, Minnesota. Association for Computational Linguistics.

\bibitem[{Dinan et~al.(2019)Dinan, Humeau, Chintagunta, and
  Weston}]{dinan-etal-2019-build}
Emily Dinan, Samuel Humeau, Bharath Chintagunta, and Jason Weston. 2019.
\newblock \href {https://doi.org/10.18653/v1/D19-1461} {Build it break it fix
  it for dialogue safety: Robustness from adversarial human attack}.
\newblock In \emph{Proceedings of the 2019 Conference on Empirical Methods in
  Natural Language Processing and the 9th International Joint Conference on
  Natural Language Processing (EMNLP-IJCNLP)}, pages 4537--4546, Hong Kong,
  China. Association for Computational Linguistics.

\bibitem[{Dua et~al.(2019)Dua, Wang, Dasigi, Stanovsky, Singh, and
  Gardner}]{dua-etal-2019-drop}
Dheeru Dua, Yizhong Wang, Pradeep Dasigi, Gabriel Stanovsky, Sameer Singh, and
  Matt Gardner. 2019.
\newblock \href {https://doi.org/10.18653/v1/N19-1246} {{DROP}: A reading
  comprehension benchmark requiring discrete reasoning over paragraphs}.
\newblock In \emph{Proceedings of the 2019 Conference of the North {A}merican
  Chapter of the Association for Computational Linguistics: Human Language
  Technologies, Volume 1 (Long and Short Papers)}, pages 2368--2378,
  Minneapolis, Minnesota. Association for Computational Linguistics.

\bibitem[{Ebrahimi et~al.(2018{\natexlab{a}})Ebrahimi, Lowd, and
  Dou}]{ebrahimi-etal-2018-adversarial}
Javid Ebrahimi, Daniel Lowd, and Dejing Dou. 2018{\natexlab{a}}.
\newblock \href {https://www.aclweb.org/anthology/C18-1055} {On adversarial
  examples for character-level neural machine translation}.
\newblock In \emph{Proceedings of the 27th International Conference on
  Computational Linguistics}, pages 653--663, Santa Fe, New Mexico, USA.
  Association for Computational Linguistics.

\bibitem[{Ebrahimi et~al.(2018{\natexlab{b}})Ebrahimi, Rao, Lowd, and
  Dou}]{ebrahimi2018hotflip}
Javid Ebrahimi, Anyi Rao, Daniel Lowd, and Dejing Dou. 2018{\natexlab{b}}.
\newblock \href {https://doi.org/10.18653/v1/P18-2006} {{H}ot{F}lip: White-box
  adversarial examples for text classification}.
\newblock In \emph{Proceedings of the 56th Annual Meeting of the Association
  for Computational Linguistics (Volume 2: Short Papers)}, pages 31--36,
  Melbourne, Australia. Association for Computational Linguistics.

\bibitem[{Ethayarajh and Jurafsky(2020)}]{ethayarajh-jurafsky-2020-utility}
Kawin Ethayarajh and Dan Jurafsky. 2020.
\newblock \href {https://www.aclweb.org/anthology/2020.emnlp-main.393} {Utility
  is in the eye of the user: A critique of {NLP} leaderboard design}.
\newblock In \emph{Proceedings of the 2020 Conference on Empirical Methods in
  Natural Language Processing (EMNLP)}, pages 4846--4853, Online. Association
  for Computational Linguistics.

\bibitem[{Ettinger(2020)}]{ettinger-2020-bert}
Allyson Ettinger. 2020.
\newblock \href {https://doi.org/10.1162/tacl_a_00298} {What {BERT} is not:
  Lessons from a new suite of psycholinguistic diagnostics for language
  models}.
\newblock \emph{Transactions of the Association for Computational Linguistics},
  8:34--48.

\bibitem[{Ettinger et~al.(2017)Ettinger, Rao, Daum{\'e}~III, and
  Bender}]{ettinger-etal-2017-towards}
Allyson Ettinger, Sudha Rao, Hal Daum{\'e}~III, and Emily~M. Bender. 2017.
\newblock \href {https://doi.org/10.18653/v1/W17-5401} {Towards linguistically
  generalizable {NLP} systems: A workshop and shared task}.
\newblock In \emph{Proceedings of the First Workshop on Building Linguistically
  Generalizable {NLP} Systems}, pages 1--10, Copenhagen, Denmark. Association
  for Computational Linguistics.

\bibitem[{Gardner et~al.(2020)Gardner, Artzi, Basmov, Berant, Bogin, Chen,
  Dasigi, Dua, Elazar, Gottumukkala, Gupta, Hajishirzi, Ilharco, Khashabi, Lin,
  Liu, Liu, Mulcaire, Ning, Singh, Smith, Subramanian, Tsarfaty, Wallace,
  Zhang, and Zhou}]{gardner-etal-2020-evaluating}
Matt Gardner, Yoav Artzi, Victoria Basmov, Jonathan Berant, Ben Bogin, Sihao
  Chen, Pradeep Dasigi, Dheeru Dua, Yanai Elazar, Ananth Gottumukkala, Nitish
  Gupta, Hannaneh Hajishirzi, Gabriel Ilharco, Daniel Khashabi, Kevin Lin,
  Jiangming Liu, Nelson~F. Liu, Phoebe Mulcaire, Qiang Ning, Sameer Singh,
  Noah~A. Smith, Sanjay Subramanian, Reut Tsarfaty, Eric Wallace, Ally Zhang,
  and Ben Zhou. 2020.
\newblock \href {https://www.aclweb.org/anthology/2020.findings-emnlp.117}
  {Evaluating models{'} local decision boundaries via contrast sets}.
\newblock In \emph{Proceedings of the 2020 Conference on Empirical Methods in
  Natural Language Processing: Findings}, pages 1307--1323, Online. Association
  for Computational Linguistics.

\bibitem[{Gauthier et~al.(2020)Gauthier, Hu, Wilcox, Qian, and
  Levy}]{gauthier-etal-2020-syntaxgym}
Jon Gauthier, Jennifer Hu, Ethan Wilcox, Peng Qian, and Roger Levy. 2020.
\newblock \href {https://doi.org/10.18653/v1/2020.acl-demos.10} {{S}yntax{G}ym:
  An online platform for targeted evaluation of language models}.
\newblock In \emph{Proceedings of the 58th Annual Meeting of the Association
  for Computational Linguistics: System Demonstrations}, pages 70--76, Online.
  Association for Computational Linguistics.

\bibitem[{Geiger et~al.(2018)Geiger, Cases, Karttunen, and
  Potts}]{geiger2018stress}
Atticus Geiger, Ignacio Cases, Lauri Karttunen, and Christopher Potts. 2018.
\newblock \href {https://arxiv.org/abs/1810.13033} {Stress-testing neural
  models of natural language inference with multiply-quantified sentences}.
\newblock \emph{arXiv preprint arXiv:1810.13033}.

\bibitem[{Glockner et~al.(2018)Glockner, Shwartz, and
  Goldberg}]{glockner-etal-2018-breaking}
Max Glockner, Vered Shwartz, and Yoav Goldberg. 2018.
\newblock \href {https://doi.org/10.18653/v1/P18-2103} {Breaking {NLI} systems
  with sentences that require simple lexical inferences}.
\newblock In \emph{Proceedings of the 56th Annual Meeting of the Association
  for Computational Linguistics (Volume 2: Short Papers)}, pages 650--655,
  Melbourne, Australia. Association for Computational Linguistics.

\bibitem[{Gururangan et~al.(2018)Gururangan, Swayamdipta, Levy, Schwartz,
  Bowman, and Smith}]{gururangan-etal-2018-annotation}
Suchin Gururangan, Swabha Swayamdipta, Omer Levy, Roy Schwartz, Samuel Bowman,
  and Noah~A. Smith. 2018.
\newblock \href {https://doi.org/10.18653/v1/N18-2017} {Annotation artifacts in
  natural language inference data}.
\newblock In \emph{Proceedings of the 2018 Conference of the North {A}merican
  Chapter of the Association for Computational Linguistics: Human Language
  Technologies, Volume 2 (Short Papers)}, pages 107--112, New Orleans,
  Louisiana. Association for Computational Linguistics.

\bibitem[{Hancock et~al.(2019)Hancock, Bordes, Mazare, and
  Weston}]{hancock-etal-2019-learning}
Braden Hancock, Antoine Bordes, Pierre-Emmanuel Mazare, and Jason Weston. 2019.
\newblock \href {https://doi.org/10.18653/v1/P19-1358} {Learning from dialogue
  after deployment: Feed yourself, chatbot!}
\newblock In \emph{Proceedings of the 57th Annual Meeting of the Association
  for Computational Linguistics}, pages 3667--3684, Florence, Italy.
  Association for Computational Linguistics.

\bibitem[{Hossain et~al.(2020)Hossain, Kovatchev, Dutta, Kao, Wei, and
  Blanco}]{hossain-etal-2020-analysis}
Md~Mosharaf Hossain, Venelin Kovatchev, Pranoy Dutta, Tiffany Kao, Elizabeth
  Wei, and Eduardo Blanco. 2020.
\newblock \href {https://www.aclweb.org/anthology/2020.emnlp-main.732} {An
  analysis of natural language inference benchmarks through the lens of
  negation}.
\newblock In \emph{Proceedings of the 2020 Conference on Empirical Methods in
  Natural Language Processing (EMNLP)}, pages 9106--9118, Online. Association
  for Computational Linguistics.

\bibitem[{Howard and Ruder(2018)}]{howard-ruder-2018-universal}
Jeremy Howard and Sebastian Ruder. 2018.
\newblock \href {https://doi.org/10.18653/v1/P18-1031} {Universal language
  model fine-tuning for text classification}.
\newblock In \emph{Proceedings of the 56th Annual Meeting of the Association
  for Computational Linguistics (Volume 1: Long Papers)}, pages 328--339,
  Melbourne, Australia. Association for Computational Linguistics.

\bibitem[{Huang et~al.(2020)Huang, Liu, and Bowman}]{huang2020counterfactually}
William Huang, Haokun Liu, and Samuel~R Bowman. 2020.
\newblock Counterfactually-augmented snli training data does not yield better
  generalization than unaugmented data.
\newblock \emph{arXiv preprint arXiv:2010.04762}.

\bibitem[{Jeretic et~al.(2020)Jeretic, Warstadt, Bhooshan, and
  Williams}]{jeretic-etal-2020-natural}
Paloma Jeretic, Alex Warstadt, Suvrat Bhooshan, and Adina Williams. 2020.
\newblock \href {https://doi.org/10.18653/v1/2020.acl-main.768} {Are natural
  language inference models {IMPPRESsive}? {L}earning {IMPlicature} and
  {PRESupposition}}.
\newblock In \emph{Proceedings of the 58th Annual Meeting of the Association
  for Computational Linguistics}, pages 8690--8705, Online. Association for
  Computational Linguistics.

\bibitem[{Jia and Liang(2017)}]{jia2017adversarial}
Robin Jia and Percy Liang. 2017.
\newblock \href {https://doi.org/10.18653/v1/D17-1215} {Adversarial examples
  for evaluating reading comprehension systems}.
\newblock In \emph{Proceedings of the 2017 Conference on Empirical Methods in
  Natural Language Processing}, pages 2021--2031, Copenhagen, Denmark.
  Association for Computational Linguistics.

\bibitem[{Kaushik et~al.(2020)Kaushik, Hovy, and Lipton}]{kaushik2019learning}
Divyansh Kaushik, Eduard Hovy, and Zachary~C Lipton. 2020.
\newblock \href {https://arxiv.org/abs/1909.12434} {Learning the difference
  that makes a difference with counterfactually-augmented data}.
\newblock \emph{International Conference on Learning Representations (ICLR)}.

\bibitem[{Khayrallah and Koehn(2018)}]{khayrallah-koehn-2018-impact}
Huda Khayrallah and Philipp Koehn. 2018.
\newblock \href {https://doi.org/10.18653/v1/W18-2709} {On the impact of
  various types of noise on neural machine translation}.
\newblock In \emph{Proceedings of the 2nd Workshop on Neural Machine
  Translation and Generation}, pages 74--83, Melbourne, Australia. Association
  for Computational Linguistics.

\bibitem[{Kim and Linzen(2020)}]{kim-linzen-2020-cogs}
Najoung Kim and Tal Linzen. 2020.
\newblock \href {https://www.aclweb.org/anthology/2020.emnlp-main.731} {{COGS}:
  A compositional generalization challenge based on semantic interpretation}.
\newblock In \emph{Proceedings of the 2020 Conference on Empirical Methods in
  Natural Language Processing (EMNLP)}, pages 9087--9105, Online. Association
  for Computational Linguistics.

\bibitem[{Kwiatkowski et~al.(2019)Kwiatkowski, Palomaki, Redfield, Collins,
  Parikh, Alberti, Epstein, Polosukhin, Devlin, Lee, Toutanova, Jones, Kelcey,
  Chang, Dai, Uszkoreit, Le, and Petrov}]{kwiatkowski-etal-2019-natural}
Tom Kwiatkowski, Jennimaria Palomaki, Olivia Redfield, Michael Collins, Ankur
  Parikh, Chris Alberti, Danielle Epstein, Illia Polosukhin, Jacob Devlin,
  Kenton Lee, Kristina Toutanova, Llion Jones, Matthew Kelcey, Ming-Wei Chang,
  Andrew~M. Dai, Jakob Uszkoreit, Quoc Le, and Slav Petrov. 2019.
\newblock \href {https://doi.org/10.1162/tacl_a_00276} {Natural questions: A
  benchmark for question answering research}.
\newblock \emph{Transactions of the Association for Computational Linguistics},
  7:452--466.

\bibitem[{{Leader Maynard} and Benesch(2016)}]{LeaderMaynard2016}
Jonathan {Leader Maynard} and Susan Benesch. 2016.
\newblock \href {https://doi.org/10.5038/1911-9933.9.3.1317} {{Dangerous Speech
  and Dangerous Ideology: An Integrated Model for Monitoring and Prevention}}.
\newblock \emph{Genocide Studies and Prevention}, 9(3):70--95.

\bibitem[{Lewis et~al.(2020)Lewis, Stenetorp, and Riedel}]{lewis2020question}
Patrick Lewis, Pontus Stenetorp, and Sebastian Riedel. 2020.
\newblock Question and answer test-train overlap in open-domain question
  answering datasets.
\newblock \emph{arXiv preprint arXiv:2008.02637}.

\bibitem[{Linzen(2020)}]{linzen-2020-accelerate}
Tal Linzen. 2020.
\newblock \href {https://doi.org/10.18653/v1/2020.acl-main.465} {How can we
  accelerate progress towards human-like linguistic generalization?}
\newblock In \emph{Proceedings of the 58th Annual Meeting of the Association
  for Computational Linguistics}, pages 5210--5217, Online. Association for
  Computational Linguistics.

\bibitem[{Liu et~al.(2003)Liu, Lieberman, and
  Selker}]{Liu+Lieberman+Selker:03a}
Hugo Liu, Henry Lieberman, and Ted Selker. 2003.
\newblock \href
  {https://web.media.mit.edu/~lieber/Publications/Affec_Sensing.pdf} {A model
  of textual affect sensing using real-world knowledge}.
\newblock In \emph{Proceedings of Intelligent User Interfaces (IUI)}, pages
  125--132.

\bibitem[{Liu et~al.(2019{\natexlab{a}})Liu, Schwartz, and
  Smith}]{liu-etal-2019-inoculation}
Nelson~F. Liu, Roy Schwartz, and Noah~A. Smith. 2019{\natexlab{a}}.
\newblock \href {https://doi.org/10.18653/v1/N19-1225} {Inoculation by
  fine-tuning: A method for analyzing challenge datasets}.
\newblock In \emph{Proceedings of the 2019 Conference of the North {A}merican
  Chapter of the Association for Computational Linguistics: Human Language
  Technologies, Volume 1 (Long and Short Papers)}, pages 2171--2179,
  Minneapolis, Minnesota. Association for Computational Linguistics.

\bibitem[{Liu et~al.(2019{\natexlab{b}})Liu, Ott, Goyal, Du, Joshi, Chen, Levy,
  Lewis, Zettlemoyer, and Stoyanov}]{liu2019roberta}
Yinhan Liu, Myle Ott, Naman Goyal, Jingfei Du, Mandar Joshi, Danqi Chen, Omer
  Levy, Mike Lewis, Luke Zettlemoyer, and Veselin Stoyanov. 2019{\natexlab{b}}.
\newblock \href {http://arxiv.org/abs/1907.11692} {{RoBERTa}: {A} robustly
  optimized {BERT} pretraining approach}.
\newblock \emph{CoRR}, abs/1907.11692.

\bibitem[{Luong et~al.(2015)Luong, Pham, and
  Manning}]{luong-etal-2015-effective}
Thang Luong, Hieu Pham, and Christopher~D. Manning. 2015.
\newblock \href {https://doi.org/10.18653/v1/D15-1166} {Effective approaches to
  attention-based neural machine translation}.
\newblock In \emph{Proceedings of the 2015 Conference on Empirical Methods in
  Natural Language Processing}, pages 1412--1421, Lisbon, Portugal. Association
  for Computational Linguistics.

\bibitem[{Marcus et~al.(1993)Marcus, Santorini, and
  Marcinkiewicz}]{marcus-etal-1993-building}
Mitchell~P. Marcus, Beatrice Santorini, and Mary~Ann Marcinkiewicz. 1993.
\newblock \href {https://www.aclweb.org/anthology/J93-2004} {Building a large
  annotated corpus of {E}nglish: The {P}enn {T}reebank}.
\newblock \emph{Computational Linguistics}, 19(2):313--330.

\bibitem[{May et~al.(2019)May, Wang, Bordia, Bowman, and
  Rudinger}]{may-etal-2019-measuring}
Chandler May, Alex Wang, Shikha Bordia, Samuel~R. Bowman, and Rachel Rudinger.
  2019.
\newblock \href {https://doi.org/10.18653/v1/N19-1063} {On measuring social
  biases in sentence encoders}.
\newblock In \emph{Proceedings of the 2019 Conference of the North {A}merican
  Chapter of the Association for Computational Linguistics: Human Language
  Technologies, Volume 1 (Long and Short Papers)}, pages 622--628, Minneapolis,
  Minnesota. Association for Computational Linguistics.

\bibitem[{McCann et~al.(2018)McCann, Keskar, Xiong, and
  Socher}]{McCann2018decaNLP}
Bryan McCann, Nitish~Shirish Keskar, Caiming Xiong, and Richard Socher. 2018.
\newblock \href {https://arxiv.org/abs/1806.08730} {The natural language
  decathlon: Multitask learning as question answering}.
\newblock \emph{arXiv preprint arXiv:1806.08730}.

\bibitem[{McCoy et~al.(2019)McCoy, Pavlick, and Linzen}]{mccoy-etal-2019-right}
Tom McCoy, Ellie Pavlick, and Tal Linzen. 2019.
\newblock \href {https://doi.org/10.18653/v1/P19-1334} {Right for the wrong
  reasons: Diagnosing syntactic heuristics in natural language inference}.
\newblock In \emph{Proceedings of the 57th Annual Meeting of the Association
  for Computational Linguistics}, pages 3428--3448, Florence, Italy.
  Association for Computational Linguistics.

\bibitem[{Minervini and Riedel(2018)}]{minervini-riedel-2018-adversarially}
Pasquale Minervini and Sebastian Riedel. 2018.
\newblock \href {https://doi.org/10.18653/v1/K18-1007} {Adversarially
  regularising neural {NLI} models to integrate logical background knowledge}.
\newblock In \emph{Proceedings of the 22nd Conference on Computational Natural
  Language Learning}, pages 65--74, Brussels, Belgium. Association for
  Computational Linguistics.

\bibitem[{Mitchell et~al.(2018)Mitchell, Cohen, Hruschka, Talukdar, Yang,
  Betteridge, Carlson, Dalvi, Gardner, Kisiel et~al.}]{mitchell2018never}
Tom Mitchell, William Cohen, Estevam Hruschka, Partha Talukdar, Bishan Yang,
  Justin Betteridge, Andrew Carlson, Bhavana Dalvi, Matt Gardner, Bryan Kisiel,
  et~al. 2018.
\newblock \href {https://dl.acm.org/doi/10.1145/3191513} {Never-ending
  learning}.
\newblock \emph{Communications of the ACM}, 61(5):103--115.

\bibitem[{Munro et~al.(2010)Munro, Bethard, Kuperman, Lai, Melnick, Potts,
  Schnoebelen, and Tily}]{munro-etal-2010-crowdsourcing}
Robert Munro, Steven Bethard, Victor Kuperman, Vicky~Tzuyin Lai, Robin Melnick,
  Christopher Potts, Tyler Schnoebelen, and Harry Tily. 2010.
\newblock \href {https://www.aclweb.org/anthology/W10-0719} {Crowdsourcing and
  language studies: the new generation of linguistic data}.
\newblock In \emph{Proceedings of the {NAACL} {HLT} 2010 Workshop on Creating
  Speech and Language Data with {A}mazon{'}s Mechanical Turk}, pages 122--130,
  Los Angeles. Association for Computational Linguistics.

\bibitem[{Naik et~al.(2018)Naik, Ravichander, Sadeh, Rose, and
  Neubig}]{naik-etal-2018-stress}
Aakanksha Naik, Abhilasha Ravichander, Norman Sadeh, Carolyn Rose, and Graham
  Neubig. 2018.
\newblock \href {https://www.aclweb.org/anthology/C18-1198} {Stress test
  evaluation for natural language inference}.
\newblock In \emph{Proceedings of the 27th International Conference on
  Computational Linguistics}, pages 2340--2353, Santa Fe, New Mexico, USA.
  Association for Computational Linguistics.

\bibitem[{Nangia et~al.(2020)Nangia, Vania, Bhalerao, and
  Bowman}]{nangia-etal-2020-crows}
Nikita Nangia, Clara Vania, Rasika Bhalerao, and Samuel~R. Bowman. 2020.
\newblock \href {https://www.aclweb.org/anthology/2020.emnlp-main.154}
  {{C}row{S}-pairs: A challenge dataset for measuring social biases in masked
  language models}.
\newblock In \emph{Proceedings of the 2020 Conference on Empirical Methods in
  Natural Language Processing (EMNLP)}, pages 1953--1967, Online. Association
  for Computational Linguistics.

\bibitem[{Neviarouskaya et~al.(2010)Neviarouskaya, Prendinger, and
  Ishizuka}]{neviarouskaya-etal-2010-recognition}
Alena Neviarouskaya, Helmut Prendinger, and Mitsuru Ishizuka. 2010.
\newblock \href {https://www.aclweb.org/anthology/C10-1091} {Recognition of
  affect, judgment, and appreciation in text}.
\newblock In \emph{Proceedings of the 23rd International Conference on
  Computational Linguistics (Coling 2010)}, pages 806--814, Beijing, China.
  Coling 2010 Organizing Committee.

\bibitem[{Nie et~al.(2019)Nie, Wang, and Bansal}]{Nie_Wang_Bansal_2019}
Yixin Nie, Yicheng Wang, and Mohit Bansal. 2019.
\newblock \href {https://doi.org/10.1609/aaai.v33i01.33016867} {Analyzing
  compositionality-sensitivity of nli models}.
\newblock \emph{Proceedings of the AAAI Conference on Artificial Intelligence},
  33(01):6867--6874.

\bibitem[{Nie et~al.(2020)Nie, Williams, Dinan, Bansal, Weston, and
  Kiela}]{nie-etal-2020-adversarial}
Yixin Nie, Adina Williams, Emily Dinan, Mohit Bansal, Jason Weston, and Douwe
  Kiela. 2020.
\newblock \href {https://doi.org/10.18653/v1/2020.acl-main.441} {Adversarial
  {NLI}: A new benchmark for natural language understanding}.
\newblock In \emph{Proceedings of the 58th Annual Meeting of the Association
  for Computational Linguistics}, pages 4885--4901, Online. Association for
  Computational Linguistics.

\bibitem[{Pang and Lee(2008)}]{PangLee08}
Bo~Pang and Lillian Lee. 2008.
\newblock \href {http://www.cs.cornell.edu/home/llee/omsa/omsa.pdf} {Opinion
  mining and sentiment analysis}.
\newblock \emph{Foundations and Trends in Information Retrieval}, 2(1):1--135.

\bibitem[{Paszke et~al.(2019)Paszke, Gross, Massa, Lerer, Bradbury, Chanan,
  Killeen, Lin, Gimelshein, Antiga, Desmaison, Kopf, Yang, DeVito, Raison,
  Tejani, Chilamkurthy, Steiner, Fang, Bai, and Chintala}]{Pytorch:2019neurips}
Adam Paszke, Sam Gross, Francisco Massa, Adam Lerer, James Bradbury, Gregory
  Chanan, Trevor Killeen, Zeming Lin, Natalia Gimelshein, Luca Antiga, Alban
  Desmaison, Andreas Kopf, Edward Yang, Zachary DeVito, Martin Raison, Alykhan
  Tejani, Sasank Chilamkurthy, Benoit Steiner, Lu~Fang, Junjie Bai, and Soumith
  Chintala. 2019.
\newblock \href
  {https://proceedings.neurips.cc/paper/2019/file/bdbca288fee7f92f2bfa9f7012727740-Paper.pdf}
  {Pytorch: An imperative style, high-performance deep learning library}.
\newblock In \emph{Advances in Neural Information Processing Systems},
  volume~32, pages 8026--8037. Curran Associates, Inc.

\bibitem[{Poletto et~al.(2020)Poletto, Basile, Sanguinetti, Bosco, and
  Patti}]{Poletto2020}
Fabio Poletto, Valerio Basile, Manuela Sanguinetti, Cristina Bosco, and Viviana
  Patti. 2020.
\newblock \href {https://doi.org/10.1007/s10579-020-09502-8} {{Resources and
  benchmark corpora for hate speech detection: a systematic review}}.
\newblock \emph{Language Resources and Evaluation}.

\bibitem[{Poliak et~al.(2018)Poliak, Naradowsky, Haldar, Rudinger, and
  Van~Durme}]{poliak-etal-2018-hypothesis}
Adam Poliak, Jason Naradowsky, Aparajita Haldar, Rachel Rudinger, and Benjamin
  Van~Durme. 2018.
\newblock \href {https://doi.org/10.18653/v1/S18-2023} {Hypothesis only
  baselines in natural language inference}.
\newblock In \emph{Proceedings of the Seventh Joint Conference on Lexical and
  Computational Semantics}, pages 180--191, New Orleans, Louisiana. Association
  for Computational Linguistics.

\bibitem[{Potts et~al.(2020)Potts, Wu, Geiger, and
  Kiela}]{potts-etal-2020-dynasent}
Christopher Potts, Zhengxuan Wu, Atticus Geiger, and Douwe Kiela. 2020.
\newblock \href {https://arxiv.org/abs/2012.15349} {{DynaSent}: A dynamic
  benchmark for sentiment analysis}.
\newblock \emph{arXiv preprint arXiv:2012.15349}.

\bibitem[{Pradhan et~al.(2012)Pradhan, Moschitti, Xue, Uryupina, and
  Zhang}]{pradhan-etal-2012-conll}
Sameer Pradhan, Alessandro Moschitti, Nianwen Xue, Olga Uryupina, and Yuchen
  Zhang. 2012.
\newblock \href {https://www.aclweb.org/anthology/W12-4501} {{C}o{NLL}-2012
  shared task: Modeling multilingual unrestricted coreference in
  {O}nto{N}otes}.
\newblock In \emph{Joint Conference on {EMNLP} and {C}o{NLL} - Shared Task},
  pages 1--40, Jeju Island, Korea. Association for Computational Linguistics.

\bibitem[{Radford et~al.(2019)Radford, Wu, Child, Luan, Amodei, and
  Sutskever}]{radford2019gpt2}
Alec Radford, Jeffrey Wu, Rewon Child, David Luan, Dario Amodei, and Ilya
  Sutskever. 2019.
\newblock \href {http://www.persagen.com/files/misc/radford2019language.pdf}
  {Language models are unsupervised multitask learners}.
\newblock \emph{OpenAI blog}, 1(8):9.

\bibitem[{Rajpurkar et~al.(2018)Rajpurkar, Jia, and
  Liang}]{rajpurkar-etal-2018-know}
Pranav Rajpurkar, Robin Jia, and Percy Liang. 2018.
\newblock \href {https://doi.org/10.18653/v1/P18-2124} {Know what you don{'}t
  know: Unanswerable questions for {SQ}u{AD}}.
\newblock In \emph{Proceedings of the 56th Annual Meeting of the Association
  for Computational Linguistics (Volume 2: Short Papers)}, pages 784--789,
  Melbourne, Australia. Association for Computational Linguistics.

\bibitem[{Rajpurkar et~al.(2016)Rajpurkar, Zhang, Lopyrev, and
  Liang}]{rajpurkar2016squad}
Pranav Rajpurkar, Jian Zhang, Konstantin Lopyrev, and Percy Liang. 2016.
\newblock \href {https://doi.org/10.18653/v1/D16-1264} {{SQ}u{AD}: 100,000+
  questions for machine comprehension of text}.
\newblock In \emph{Proceedings of the 2016 Conference on Empirical Methods in
  Natural Language Processing}, pages 2383--2392, Austin, Texas. Association
  for Computational Linguistics.

\bibitem[{Ribeiro et~al.(2020)Ribeiro, Wu, Guestrin, and
  Singh}]{ribeiro-etal-2020-beyond}
Marco~Tulio Ribeiro, Tongshuang Wu, Carlos Guestrin, and Sameer Singh. 2020.
\newblock \href {https://doi.org/10.18653/v1/2020.acl-main.442} {Beyond
  accuracy: Behavioral testing of {NLP} models with {C}heck{L}ist}.
\newblock In \emph{Proceedings of the 58th Annual Meeting of the Association
  for Computational Linguistics}, pages 4902--4912, Online. Association for
  Computational Linguistics.

\bibitem[{Rozen et~al.(2019)Rozen, Shwartz, Aharoni, and
  Dagan}]{rozen-etal-2019-diversify}
Ohad Rozen, Vered Shwartz, Roee Aharoni, and Ido Dagan. 2019.
\newblock \href {https://doi.org/10.18653/v1/K19-1019} {Diversify your
  datasets: Analyzing generalization via controlled variance in adversarial
  datasets}.
\newblock In \emph{Proceedings of the 23rd Conference on Computational Natural
  Language Learning (CoNLL)}, pages 196--205, Hong Kong, China. Association for
  Computational Linguistics.

\bibitem[{Rudinger et~al.(2018)Rudinger, Naradowsky, Leonard, and
  Van~Durme}]{rudinger-etal-2018-gender}
Rachel Rudinger, Jason Naradowsky, Brian Leonard, and Benjamin Van~Durme. 2018.
\newblock \href {https://doi.org/10.18653/v1/N18-2002} {Gender bias in
  coreference resolution}.
\newblock In \emph{Proceedings of the 2018 Conference of the North {A}merican
  Chapter of the Association for Computational Linguistics: Human Language
  Technologies, Volume 2 (Short Papers)}, pages 8--14, New Orleans, Louisiana.
  Association for Computational Linguistics.

\bibitem[{Röttger et~al.(2020)Röttger, Vidgen, Nguyen, Waseem, Margetts, and
  Pierrehumbert}]{rottger2020hatecheck}
Paul Röttger, Bertram Vidgen, Dong Nguyen, Zeerak Waseem, Helen Margetts, and
  Janet Pierrehumbert. 2020.
\newblock \href {http://arxiv.org/abs/2012.15606} {Hatecheck: Functional tests
  for hate speech detection models}.

\bibitem[{Saha et~al.(2020)Saha, Nie, and Bansal}]{saha-etal-2020-conjnli}
Swarnadeep Saha, Yixin Nie, and Mohit Bansal. 2020.
\newblock \href {https://www.aclweb.org/anthology/2020.emnlp-main.661}
  {{C}onj{NLI}: Natural language inference over conjunctive sentences}.
\newblock In \emph{Proceedings of the 2020 Conference on Empirical Methods in
  Natural Language Processing (EMNLP)}, pages 8240--8252, Online. Association
  for Computational Linguistics.

\bibitem[{S{\'a}nchez~Valencia(1991)}]{sanchez1991studies}
Victor S{\'a}nchez~Valencia. 1991.
\newblock \href
  {https://www.illc.uva.nl/Research/Publications/Dissertations/HDS-17-Victor-Sanchez.text.pdf}
  {\emph{Studies on natural logic and categorial grammar, University of
  Amsterdam Ph. D}}.
\newblock Ph.D. thesis, thesis.

\bibitem[{Sankar et~al.(2019)Sankar, Subramanian, Pal, Chandar, and
  Bengio}]{sankar-etal-2019-neural}
Chinnadhurai Sankar, Sandeep Subramanian, Chris Pal, Sarath Chandar, and Yoshua
  Bengio. 2019.
\newblock \href {https://doi.org/10.18653/v1/P19-1004} {Do neural dialog
  systems use the conversation history effectively? an empirical study}.
\newblock In \emph{Proceedings of the 57th Annual Meeting of the Association
  for Computational Linguistics}, pages 32--37, Florence, Italy. Association
  for Computational Linguistics.

\bibitem[{Schuster et~al.(2020)Schuster, Chen, and
  Degen}]{schuster-etal-2020-harnessing}
Sebastian Schuster, Yuxing Chen, and Judith Degen. 2020.
\newblock \href {https://doi.org/10.18653/v1/2020.acl-main.479} {Harnessing the
  linguistic signal to predict scalar inferences}.
\newblock In \emph{Proceedings of the 58th Annual Meeting of the Association
  for Computational Linguistics}, pages 5387--5403, Online. Association for
  Computational Linguistics.

\bibitem[{Seo et~al.(2017)Seo, Kembhavi, Farhadi, and
  Hajishirzi}]{Seo2016BidAF}
Minjoon Seo, Aniruddha Kembhavi, Ali Farhadi, and Hannaneh Hajishirzi. 2017.
\newblock \href {https://openreview.net/forum?id=HJ0UKP9ge} {Bidirectional
  attention flow for machine comprehension}.
\newblock In \emph{The International Conference on Learning Representations
  (ICLR)}.

\bibitem[{Shuster et~al.(2020)Shuster, Ju, Roller, Dinan, Boureau, and
  Weston}]{shuster-etal-2020-dialogue}
Kurt Shuster, Da~Ju, Stephen Roller, Emily Dinan, Y-Lan Boureau, and Jason
  Weston. 2020.
\newblock \href {https://doi.org/10.18653/v1/2020.acl-main.222} {The dialogue
  dodecathlon: Open-domain knowledge and image grounded conversational agents}.
\newblock In \emph{Proceedings of the 58th Annual Meeting of the Association
  for Computational Linguistics}, pages 2453--2470, Online. Association for
  Computational Linguistics.

\bibitem[{Silver et~al.(2013)Silver, Yang, and Li}]{silver2013lifelong}
Daniel~L Silver, Qiang Yang, and Lianghao Li. 2013.
\newblock \href
  {http://axon.cs.byu.edu/~martinez/classes/678/Presentations/Martin.pdf}
  {Lifelong machine learning systems: Beyond learning algorithms}.
\newblock In \emph{2013 AAAI spring symposium series}.

\bibitem[{Snow et~al.(2008)Snow, O{'}Connor, Jurafsky, and
  Ng}]{snow-etal-2008-cheap}
Rion Snow, Brendan O{'}Connor, Daniel Jurafsky, and Andrew Ng. 2008.
\newblock \href {https://www.aclweb.org/anthology/D08-1027} {Cheap and fast
  {--} but is it good? evaluating non-expert annotations for natural language
  tasks}.
\newblock In \emph{Proceedings of the 2008 Conference on Empirical Methods in
  Natural Language Processing}, pages 254--263, Honolulu, Hawaii. Association
  for Computational Linguistics.

\bibitem[{Socher et~al.(2013)Socher, Perelygin, Wu, Chuang, Manning, Ng, and
  Potts}]{socher-etal-2013-recursive}
Richard Socher, Alex Perelygin, Jean Wu, Jason Chuang, Christopher~D. Manning,
  Andrew Ng, and Christopher Potts. 2013.
\newblock \href {https://www.aclweb.org/anthology/D13-1170} {Recursive deep
  models for semantic compositionality over a sentiment treebank}.
\newblock In \emph{Proceedings of the 2013 Conference on Empirical Methods in
  Natural Language Processing}, pages 1631--1642, Seattle, Washington, USA.
  Association for Computational Linguistics.

\bibitem[{Stiennon et~al.(2020)Stiennon, Ouyang, Wu, Ziegler, Lowe, Voss,
  Radford, Amodei, and Christiano}]{stiennon2020learning}
Nisan Stiennon, Long Ouyang, Jeffrey Wu, Daniel Ziegler, Ryan Lowe, Chelsea
  Voss, Alec Radford, Dario Amodei, and Paul~F Christiano. 2020.
\newblock \href {https://arxiv.org/abs/2009.01325} {Learning to summarize with
  human feedback}.
\newblock \emph{Advances in Neural Information Processing Systems}, 33.

\bibitem[{Sudhof et~al.(2014)Sudhof, G{\'o}mez~Emilsson, Maas, and
  Potts}]{Sudhof-etal:2014}
Moritz Sudhof, Andr{\'e}s G{\'o}mez~Emilsson, Andrew~L. Maas, and Christopher
  Potts. 2014.
\newblock \href {https://doi.org/10.1145/2623330.2623687} {Sentiment expression
  conditioned by affective transitions and social forces}.
\newblock In \emph{Proceedings of 20th Conference on Knowledge Discovery and
  Data Mining}, pages 1136--1145, New York. ACM.

\bibitem[{Sugawara et~al.(2020)Sugawara, Stenetorp, Inui, and
  Aizawa}]{sugawara2020assessing}
Saku Sugawara, Pontus Stenetorp, Kentaro Inui, and Akiko Aizawa. 2020.
\newblock Assessing the benchmarking capacity of machine reading comprehension
  datasets.
\newblock In \emph{Proceedings of the AAAI Conference on Artificial
  Intelligence}, volume~34, pages 8918--8927.

\bibitem[{Sundararajan et~al.(2017)Sundararajan, Taly, and
  Yan}]{sundararajan-etal-2017-axiomatic}
Mukund Sundararajan, Ankur Taly, and Qiqi Yan. 2017.
\newblock \href {http://proceedings.mlr.press/v70/sundararajan17a.html}
  {Axiomatic attribution for deep networks}.
\newblock In \emph{Proceedings of the 34th International Conference on Machine
  Learning, {ICML} 2017, Sydney, NSW, Australia, 6-11 August 2017}, volume~70
  of \emph{Proceedings of Machine Learning Research}, pages 3319--3328. {PMLR}.

\bibitem[{Swayamdipta et~al.(2020)Swayamdipta, Schwartz, Lourie, Wang,
  Hajishirzi, Smith, and Choi}]{swayamdipta-etal-2020-dataset}
Swabha Swayamdipta, Roy Schwartz, Nicholas Lourie, Yizhong Wang, Hannaneh
  Hajishirzi, Noah~A. Smith, and Yejin Choi. 2020.
\newblock \href {https://www.aclweb.org/anthology/2020.emnlp-main.746} {Dataset
  cartography: Mapping and diagnosing datasets with training dynamics}.
\newblock In \emph{Proceedings of the 2020 Conference on Empirical Methods in
  Natural Language Processing (EMNLP)}, pages 9275--9293, Online. Association
  for Computational Linguistics.

\bibitem[{Thorne et~al.(2018)Thorne, Vlachos, Christodoulopoulos, and
  Mittal}]{thorne-etal-2018-fever}
James Thorne, Andreas Vlachos, Christos Christodoulopoulos, and Arpit Mittal.
  2018.
\newblock \href {https://doi.org/10.18653/v1/N18-1074} {{FEVER}: a large-scale
  dataset for fact extraction and {VER}ification}.
\newblock In \emph{Proceedings of the 2018 Conference of the North {A}merican
  Chapter of the Association for Computational Linguistics: Human Language
  Technologies, Volume 1 (Long Papers)}, pages 809--819, New Orleans,
  Louisiana. Association for Computational Linguistics.

\bibitem[{Tsuchiya(2018)}]{tsuchiya-2018-performance}
Masatoshi Tsuchiya. 2018.
\newblock \href {https://www.aclweb.org/anthology/L18-1239} {Performance impact
  caused by hidden bias of training data for recognizing textual entailment}.
\newblock In \emph{Proceedings of the Eleventh International Conference on
  Language Resources and Evaluation ({LREC} 2018)}, Miyazaki, Japan. European
  Language Resources Association (ELRA).

\bibitem[{van Benthem(2008)}]{vanBenthem08NATLOG}
Johan van Benthem. 2008.
\newblock A brief history of natural logic.
\newblock In \emph{Logic, Navya-Nyaya and Applications: Homage to {B}imal
  {M}atilal}.

\bibitem[{Vaswani et~al.(2017)Vaswani, Shazeer, Parmar, Uszkoreit, Jones,
  Gomez, Kaiser, and Polosukhin}]{vaswani2017attention}
Ashish Vaswani, Noam Shazeer, Niki Parmar, Jakob Uszkoreit, Llion Jones,
  Aidan~N Gomez, \L~ukasz Kaiser, and Illia Polosukhin. 2017.
\newblock \href
  {http://papers.nips.cc/paper/7181-attention-is-all-you-need.pdf} {Attention
  is all you need}.
\newblock In I.~Guyon, U.~V. Luxburg, S.~Bengio, H.~Wallach, R.~Fergus,
  S.~Vishwanathan, and R.~Garnett, editors, \emph{Advances in Neural
  Information Processing Systems 30}, pages 5998--6008. Curran Associates, Inc.

\bibitem[{Vidgen and Derczynski(2020)}]{Vidgen2020}
Bertie Vidgen and Leon Derczynski. 2020.
\newblock \href {http://arxiv.org/abs/2004.01670} {{Directions in Abusive
  Language Training Data: Garbage In, Garbage Out}}.
\newblock \emph{arXiv:2004.01670}, pages 1--26.

\bibitem[{Vidgen et~al.(2019)Vidgen, Harris, Nguyen, Tromble, Hale, and
  Margetts}]{vidgen-etal-2019-challenges}
Bertie Vidgen, Alex Harris, Dong Nguyen, Rebekah Tromble, Scott Hale, and Helen
  Margetts. 2019.
\newblock \href {https://doi.org/10.18653/v1/W19-3509} {Challenges and
  frontiers in abusive content detection}.
\newblock In \emph{Proceedings of the Third Workshop on Abusive Language
  Online}, pages 80--93, Florence, Italy. Association for Computational
  Linguistics.

\bibitem[{Vidgen et~al.(2020)Vidgen, Thrush, Waseem, and
  Kiela}]{vidgen2020learning}
Bertie Vidgen, Tristan Thrush, Zeerak Waseem, and Douwe Kiela. 2020.
\newblock \href {https://arxiv.org/abs/2012.15761} {Learning from the worst:
  Dynamically generated datasets to improve online hate detection}.
\newblock \emph{arXiv preprint arXiv:2012.15761}.

\bibitem[{Von~Ahn and Dabbish(2008)}]{vonahn2008designing}
Luis Von~Ahn and Laura Dabbish. 2008.
\newblock \href {https://dl.acm.org/doi/fullHtml/10.1145/1378704.1378719}
  {Designing games with a purpose}.
\newblock \emph{Communications of the ACM}, 51(8):58--67.

\bibitem[{Wallace et~al.(2019)Wallace, Feng, Kandpal, Gardner, and
  Singh}]{wallace-etal-2019-universal}
Eric Wallace, Shi Feng, Nikhil Kandpal, Matt Gardner, and Sameer Singh. 2019.
\newblock \href {https://doi.org/10.18653/v1/D19-1221} {Universal adversarial
  triggers for attacking and analyzing {NLP}}.
\newblock In \emph{Proceedings of the 2019 Conference on Empirical Methods in
  Natural Language Processing and the 9th International Joint Conference on
  Natural Language Processing (EMNLP-IJCNLP)}, pages 2153--2162, Hong Kong,
  China. Association for Computational Linguistics.

\bibitem[{Wang et~al.(2019)Wang, Pruksachatkun, Nangia, Singh, Michael, Hill,
  Levy, and Bowman}]{wang2019superglue}
Alex Wang, Yada Pruksachatkun, Nikita Nangia, Amanpreet Singh, Julian Michael,
  Felix Hill, Omer Levy, and Samuel Bowman. 2019.
\newblock \href
  {https://proceedings.neurips.cc/paper/2019/file/4496bf24afe7fab6f046bf4923da8de6-Paper.pdf}
  {Superglue: A stickier benchmark for general-purpose language understanding
  systems}.
\newblock In \emph{Advances in Neural Information Processing Systems},
  volume~32, pages 3266--3280. Curran Associates, Inc.

\bibitem[{Wang et~al.(2018)Wang, Singh, Michael, Hill, Levy, and
  Bowman}]{wang-etal-2018-glue}
Alex Wang, Amanpreet Singh, Julian Michael, Felix Hill, Omer Levy, and Samuel
  Bowman. 2018.
\newblock \href {https://doi.org/10.18653/v1/W18-5446} {{GLUE}: A multi-task
  benchmark and analysis platform for natural language understanding}.
\newblock In \emph{Proceedings of the 2018 {EMNLP} Workshop {B}lackbox{NLP}:
  Analyzing and Interpreting Neural Networks for {NLP}}, pages 353--355,
  Brussels, Belgium. Association for Computational Linguistics.

\bibitem[{Warstadt et~al.(2019)Warstadt, Cao, Grosu, Peng, Blix, Nie, Alsop,
  Bordia, Liu, Parrish, Wang, Phang, Mohananey, Htut, Jeretic, and
  Bowman}]{warstadt-etal-2019-investigating}
Alex Warstadt, Yu~Cao, Ioana Grosu, Wei Peng, Hagen Blix, Yining Nie, Anna
  Alsop, Shikha Bordia, Haokun Liu, Alicia Parrish, Sheng-Fu Wang, Jason Phang,
  Anhad Mohananey, Phu~Mon Htut, Paloma Jeretic, and Samuel~R. Bowman. 2019.
\newblock \href {https://doi.org/10.18653/v1/D19-1286} {Investigating
  {BERT}{'}s knowledge of language: Five analysis methods with {NPI}s}.
\newblock In \emph{Proceedings of the 2019 Conference on Empirical Methods in
  Natural Language Processing and the 9th International Joint Conference on
  Natural Language Processing (EMNLP-IJCNLP)}, pages 2877--2887, Hong Kong,
  China. Association for Computational Linguistics.

\bibitem[{Warstadt et~al.(2020)Warstadt, Parrish, Liu, Mohananey, Peng, Wang,
  and Bowman}]{warstadt-etal-2020-blimp-benchmark}
Alex Warstadt, Alicia Parrish, Haokun Liu, Anhad Mohananey, Wei Peng, Sheng-Fu
  Wang, and Samuel~R. Bowman. 2020.
\newblock \href {https://www.aclweb.org/anthology/2020.scil-1.47} {{BL}i{MP}: A
  benchmark of linguistic minimal pairs for {E}nglish}.
\newblock In \emph{Proceedings of the Society for Computation in Linguistics
  2020}, pages 409--410, New York, New York. Association for Computational
  Linguistics.

\bibitem[{Waseem et~al.(2017)Waseem, Davidson, Warmsley, and
  Weber}]{Waseem2017}
Zeerak Waseem, Thomas Davidson, Dana Warmsley, and Ingmar Weber. 2017.
\newblock \href {https://doi.org/10.1080/17421770903114687} {{Understanding
  Abuse: A Typology of Abusive Language Detection Subtasks}}.
\newblock In \emph{Proceedings of the First Workshop on Abusive Language
  Online}, pages 78--84.

\bibitem[{White et~al.(2018)White, Rudinger, Rawlins, and
  Van~Durme}]{white-etal-2018-lexicosyntactic}
Aaron~Steven White, Rachel Rudinger, Kyle Rawlins, and Benjamin Van~Durme.
  2018.
\newblock \href {https://doi.org/10.18653/v1/D18-1501} {Lexicosyntactic
  inference in neural models}.
\newblock In \emph{Proceedings of the 2018 Conference on Empirical Methods in
  Natural Language Processing}, pages 4717--4724, Brussels, Belgium.
  Association for Computational Linguistics.

\bibitem[{White et~al.(2020)White, Stengel-Eskin, Vashishtha, Govindarajan,
  Reisinger, Vieira, Sakaguchi, Zhang, Ferraro, Rudinger, Rawlins, and
  Van~Durme}]{white-etal-2020-universal}
Aaron~Steven White, Elias Stengel-Eskin, Siddharth Vashishtha,
  Venkata~Subrahmanyan Govindarajan, Dee~Ann Reisinger, Tim Vieira, Keisuke
  Sakaguchi, Sheng Zhang, Francis Ferraro, Rachel Rudinger, Kyle Rawlins, and
  Benjamin Van~Durme. 2020.
\newblock \href {https://www.aclweb.org/anthology/2020.lrec-1.699} {The
  universal decompositional semantics dataset and decomp toolkit}.
\newblock In \emph{Proceedings of the 12th Language Resources and Evaluation
  Conference}, pages 5698--5707, Marseille, France. European Language Resources
  Association.

\bibitem[{Wiebe et~al.(2005)Wiebe, Wilson, and
  Cardie}]{Wiebe:Wilson:Cardie:2005}
Janyce Wiebe, Theresa Wilson, and Claire Cardie. 2005.
\newblock \href
  {https://www.cs.cornell.edu/home/cardie/papers/lre05withappendix.pdf}
  {Annotating expressions of opinions and emotions in language}.
\newblock \emph{Language Resources and Evaluation}, 39(2--3):165--210.

\bibitem[{Williams et~al.(2018)Williams, Nangia, and
  Bowman}]{williams-etal-2018-broad}
Adina Williams, Nikita Nangia, and Samuel Bowman. 2018.
\newblock \href {https://doi.org/10.18653/v1/N18-1101} {A broad-coverage
  challenge corpus for sentence understanding through inference}.
\newblock In \emph{Proceedings of the 2018 Conference of the North {A}merican
  Chapter of the Association for Computational Linguistics: Human Language
  Technologies, Volume 1 (Long Papers)}, pages 1112--1122, New Orleans,
  Louisiana. Association for Computational Linguistics.

\bibitem[{Yang et~al.(2018)Yang, Qi, Zhang, Bengio, Cohen, Salakhutdinov, and
  Manning}]{yang-etal-2018-hotpotqa}
Zhilin Yang, Peng Qi, Saizheng Zhang, Yoshua Bengio, William Cohen, Ruslan
  Salakhutdinov, and Christopher~D. Manning. 2018.
\newblock \href {https://doi.org/10.18653/v1/D18-1259} {{H}otpot{QA}: A dataset
  for diverse, explainable multi-hop question answering}.
\newblock In \emph{Proceedings of the 2018 Conference on Empirical Methods in
  Natural Language Processing}, pages 2369--2380, Brussels, Belgium.
  Association for Computational Linguistics.

\bibitem[{Yang et~al.(2017)Yang, Zhang, Urbanek, Feng, Miller, Szlam, Kiela,
  and Weston}]{yang2017mastering}
Zhilin Yang, Saizheng Zhang, Jack Urbanek, Will Feng, Alexander~H Miller,
  Arthur Szlam, Douwe Kiela, and Jason Weston. 2017.
\newblock \href {https://arxiv.org/abs/1711.07950} {Mastering the dungeon:
  Grounded language learning by mechanical turker descent}.
\newblock \emph{arXiv preprint arXiv:1711.07950}.

\bibitem[{Yu and Ettinger(2020)}]{yu-ettinger-2020-assessing}
Lang Yu and Allyson Ettinger. 2020.
\newblock \href {https://www.aclweb.org/anthology/2020.emnlp-main.397}
  {Assessing phrasal representation and composition in transformers}.
\newblock In \emph{Proceedings of the 2020 Conference on Empirical Methods in
  Natural Language Processing (EMNLP)}, pages 4896--4907, Online. Association
  for Computational Linguistics.

\bibitem[{Zhou et~al.(2020)Zhou, Nie, Tan, and Bansal}]{zhou-etal-2020-curse}
Xiang Zhou, Yixin Nie, Hao Tan, and Mohit Bansal. 2020.
\newblock \href {https://www.aclweb.org/anthology/2020.emnlp-main.659} {The
  curse of performance instability in analysis datasets: Consequences, source,
  and suggestions}.
\newblock In \emph{Proceedings of the 2020 Conference on Empirical Methods in
  Natural Language Processing (EMNLP)}, pages 8215--8228, Online. Association
  for Computational Linguistics.

\end{thebibliography}
\bibliographystyle{acl_natbib}

%\appendix
%\section{Example Appendix}
%\label{sec:appendix}
%This is an appendix.

\end{document}